\pdfoutput=1

\documentclass[11pt]{article}

\usepackage[]{acl}

\usepackage{hyperref}
\usepackage{xcolor}
\definecolor{darkblue}{rgb}{0, 0, 0.5}

\usepackage{multirow}
\usepackage{amsmath}
\usepackage{amsfonts}
\usepackage{algorithmicx,algorithm}
\usepackage{amssymb}
\usepackage{float}
\usepackage{times}
\usepackage{latexsym}
\usepackage{courier}
\usepackage{booktabs}
\usepackage{subfigure}
\usepackage{indentfirst}
\usepackage{microtype}
\usepackage{graphicx}

\title{An Overview on Controllable Text Generation via Variational Auto-Encoders}
\author{Haoqin Tu, Yitong Li\\University of Chinese Academy of Sciences\\Noah's Ark Lab, Huawei\\ Huawei Technologies Ltd.\\ \texttt{tuisaac163@gmail.com}, \texttt{liyitong3@huawei.com}}
\begin{document}
\maketitle

\begin{abstract}
Recent advances in neural-based generative modeling have reignited the hopes of having computer systems capable of conversing with humans and able to understand natural language. The employment of deep neural architectures has been largely explored in a multitude of context and tasks to fulfill various user needs. On one hand, producing textual content that meets specific requirements is of priority for a model to seamlessly conduct conversations with different groups of people. On the other hand, latent variable models (LVM) such as variational auto-encoders (VAEs) as one of the most popular genres of generative models are designed to characterize the distributional pattern of textual data. Thus they are inherently capable of learning the integral textual features that are worth exploring for controllable pursuits. 

\noindent This overview gives an introduction to existing generation schemes, problems associated with text variational auto-encoders, and a review of several applications about the controllable generation that are instantiations of these general formulations,\footnote{A detailed paper list is available at \url{https://github.com/ImKeTT/CTG-latentAEs}} as well as related datasets, metrics and discussions for future researches. Hopefully, this overview will provide an overview of living questions, popular methodologies and raw thoughts for controllable language generation under the scope of variational auto-encoder.
\end{abstract}

\section{Introduction}
Recent employment of deep neural architectures in natural language processing (NLP) has been largely explored in a multitude of contexts and tasks to fulfill various user needs such as machine translation \cite{bahdanau2014neural}, text summarization \cite{rush2015neural}, dialogue \cite{serban2016building,mou2016sequence}, question answering \cite{iyyer2014neural}, etc. Especially, high capacity of deep learning models trained on large-scale datasets demonstrate unparalleled abilities to produce realistic and coherent texts, which benefits natural language generation (NLG), one of the most eye-catching domains in NLP. As a matter of fact, obtaining systems to automatically produce realistic-looking texts has been a goal pursued since the early stage of artificial intelligence \cite{meehan1977tale}. As an elementary task in NLP, NLG aims to generate authentic and plausible textual content that is realistic-looking \cite{turing2009computing}. Ideally, sentences generated from a language model (LM) not only ought to preserve the semantic and syntactic properties of real-world sentences but also should be varied in expression style out of diversity reason \cite{zhang2017adversarial}. Natural language generation is an inherently complex task, which requires abundant linguistic and domain knowledge at multiple levels, including syntax, semantics, morphology, phonology, pragmatics, and so on. 

In real life, it is easy for us to realize that the word context may carry different semantics for different readers. Therefore the generated texts should be tailored to their specific audience in terms of appropriateness of content and terminology usage \cite{paris2015user}, as well as for customized network environment and transparency reasons \cite{mayfield2019equity}. The goal of controllable text generation aims to generate coherent and grammatically correct texts whose attributes can be controlled \cite{elhadad1990constraint} and/or abide by user-defined rules which reflect the particular interests of system users \cite{garbacea2020neural}.

Controllable generation task inherently involves more complex and multi-dimensional data patterns. Latent variable models have natural advantages in handling this task because they provide a declarative language for specifying prior knowledge and structural relationships in complex datasets \cite{kim2018tutorial}. This property empowers the LVMs such as variational auto-encoder (VAE) \cite{bowman2015generating} with the ability to leverage constraints into its representative hidden space for controllable generation. However, the direct inference process of latent variable models requires an integral over the latent variable, which often has no analytic form or is time-consuming to compute with massive data instances. This is also a fundamental problem that Variational Inference (VI) methods, including VAE, aim to solve. There have been much recent and interesting work that emerged to combine the complementary strengths of variational inference in LVMs and deep learning. With the invention of compatible parameterization tricks such as reparameterization \cite{kingma2015variational} for variational inference with deep function approximators (deep inference), latent variable models become more competent to handle miscellaneous NLP tasks by absorbing advanced neural model techniques.

\section{Background}
\subsection{Notations}
We use lowercase letters in boldface (e.g., $\mathbf{x}$) to denote vectors, normal letters (e.g., $x, X$) to denote scalars and uppercase letters in boldface (e.g., $\mathbf{X}$) to denote sets. We use the symbol $\|\mathbf{X}\|$ to denote the size of a set. Calligraphic letters denote the functional space of a value set (i.e., $\mathcal{X}$). Both English letters and Greek letters are adopted. We use $p(\cdot)$ and $q(\cdot)$ to denote distributions and $f(\cdot)$ to denote a function, which are usually shortened to $p$, $q$, and $f$. Subscripts and superscripts are used to tell the different variables/distributions/functions apart.

\subsection{Controllable Text Generation}
We define the task of controllable text generation as finding a function $f$ to generate sentences that obey certain generation rules or conditions. This can be formally defined as: given a set of $n$ conditions $\mathbf{C}  = \{\mathbf{c}_i\}_{1}^n \in \mathcal{C}$, where $\mathcal{C}$ denotes the condition space. The goal of controllable generation is formalized as learning a function $f$:
\begin{equation}
    f(\mathbf{C}) = \mathbf{Z}, \mathbf{Z}\in \mathcal{Z},
\end{equation}
which aims at generating sentences $\mathbf{Z}$ from space $\mathcal{Z}$ that fulfill desired conditions $\mathbf{C}$. In general, the controlled sentence generation task can be divided into two categories according to the category in which restrictions are imposed: generation with soft constraint and hard constraint. 

To be more precise, (1) soft constraint text generation requires the generated sentences to be semantically similar to the given constraints (e.g., topic or style), rather than explicitly enforcing certain concepts or rules (e.g., keywords) to appear in the content. The mapping function $f$ mentioned above serves as a measurement to find sentences with the highest semantic similarity with given constraints. For example, given a corpus of (style, text) pairs as training data, followed by training a conditional language model to learn the linguistic relevance between (style, text) pairs and generate texts with such style, we achieve controllable text generation with soft constraints. (2) Hard constraint focuses on controlling specific tokens or textual structures (e.g., keywords, sentence length) during generation, thus being more fine-grained compared with the soft one. It indicates the compulsive inclusion of given constraints in the output texts. Hence, the function $f$ here is regarded as a binary sign on a specified controlling level (e.g., token, syntax) to eliminate the possibility of producing unqualified features on such a level. Typically, the condition of hard constraint is keywords and the controllable generation process requires the model to generate sentences with provided exact keywords embedded in generated contexts. 

However, generating text under specific lexical constraints is challenging \cite{zhang2020pointer}. As for the comparison of two types of controllable generation, soft constraint generation, on the other way around, is not capable of handling the explicit appearance of conditions on the token level but can produce authentic texts with particular styles or topics and more straightforward network designs as trade-offs. Hard constraint generative models process given conditions with higher proficiency by placing explicit restrictions on independent attribute controls, but often face several issues such as unitary syntax, semantical inconsistency \cite{wiseman2018learning,wiseman2017challenges}, requiring more sophisticated model architectures \cite{garbacea2020neural} and more training samples with annotations.

In the past few years, a large number of researchers have tried to use different methods for controllable text generation with both types of constraints. Compared with other potential language generating methods, such as generative adversarial networks (GAN) based \cite{yu2017seqgan,guo2018long}, plain recurrent neural network (RNN) based \cite{mikolov2010recurrent,graves2013generating} and Transformer based methods \cite{vaswani2017attention,dai2019transformer,devlin2018bert}, latent variable models such as variational auto-encoder are particularly suitable for attribute specified (or controllable) text generation, because the latent space geometry of these models conduct multiple views of knowledge in a given corpus (i.e., style, topic, and high-level linguistic or semantic features), being beneficial for controllable generation \cite{fang2019implicit}. Besides, the latent knowledge that originates from a variational auto-encoder can help mitigate against model misspecification \cite{kim2018tutorial}, can allow for data-efficient learning \cite{rezende2015variational,tomczak2016improving}, and can enable interesting structures to emerge through a carefully crafted generative model.

\subsection{Variational Auto-Encoders for Language Modeling}
\begin{figure*}[ht]
\centering
\includegraphics[width=0.9\linewidth]{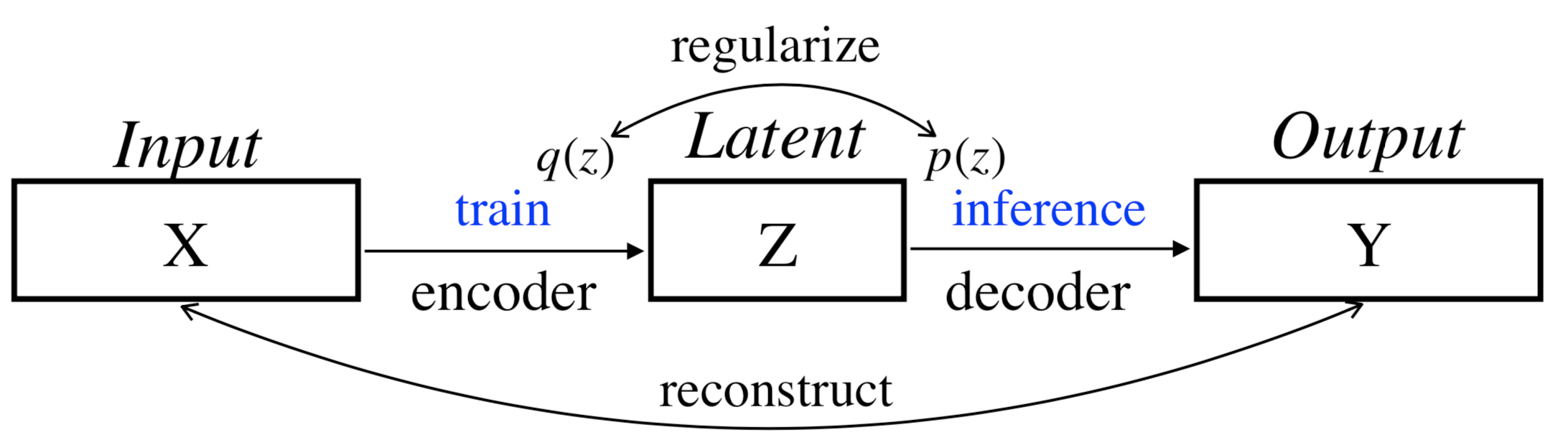}
\caption{Working scheme of a variational auto-encoder. The training loss of a variational auto-encoder essentially consists of two parts, the reconstruction loss in the observed data and the regularization loss in the latent variables.}
\label{fig:workscheme}
\end{figure*}

To model data distributions via latent variable models, we usually focus on finding the best parameter $\theta^*$ that models $\int_{\mathcal{Z}} p_{\theta^*}(\mathbf{X}, \mathbf{z})d\mathbf{z}$ to fit in the true data distribution $p(\mathbf{X})$, where observed variable set $\mathbf{X}$ with $N$ data points and latent variables $\mathbf{z}$ are considered. In practice, maximal likelihood estimation (MLE) is widely employed to set the criteria of the ``best'' distribution fit in, which aims at minimizing the average negative log-likelihood (NLL) of data $\mathbf{x}$ parameterized by $\theta$:
\begin{equation}
\begin{aligned}
& \min _{\mathbf{\theta} \in \Theta} \frac{1}{N} \sum_N-\log p_{\mathbf{\theta}}\left(\mathbf{x}\right) = \\&\min _{\mathbf{\theta} \in \Theta} \frac{1}{N} \sum_N-\int_{\mathcal{Z}}\log p_{\mathbf{\theta}}\left(\mathbf{x}, \mathbf{z}\right)d\mathbf{z},
\end{aligned}
\end{equation}
here $\mathbf{x}$ is described as observed data points. As we introduce the continuous latent variable $\mathbf{z}$ into this objective, it becomes integral intractable to be calculated directly. As an efficient alternative, amortized variational inference (AVI) takes a similar idea of estimation maximazation (EM) algorithm to iteratively optimize the parameter $\theta$ and the estimated posterior\footnote{Data distribution $p_{\theta}(\mathbf{x}) = \int_{\mathcal{Z}}p_{\mathbf{\theta}}\left(\mathbf{x}, \mathbf{z}\right)d\mathbf{z}$ and the posterior $q_{\theta}(\mathbf{z}\mid \mathbf{x}) = \frac{q_{\theta}(\mathbf{z}, \mathbf{x})}{p_{\theta}(\mathbf{x})}$ shares the equivalent integral intractability, thus the estimated posterior is generally calculated by Monte Carlo sampled latent variables.} of latent variables $q_{\theta}(\mathbf{z}\mid \mathbf{x})$. Though a neural-network-based estimator is introduced to predict function parameters more effectively, the inference of deep parameterizations in VI usually makes posterior inference intractable, and the latent variable objectives often complicate backpropagation by introducing points of non-differentiability \cite{kim2018tutorial}. In practice, variational auto-encoder widely uses the reparameterization \cite{kingma2015variational} trick during neural network back-propagation to obtain differentiable gradients with low variance. This technique requires pre-assigned continuous parametric distribution\footnote{It is usually an isotropic Gaussian with diagonal covariance matrix, however, other continuous distributions (e.g., Gamma, Dirichlet) can also be applied \cite{ruiz2016generalized,naesseth2017reparameterization}.} (or posterior distribution) $q_{\phi}(\mathbf{z}\mid \mathbf{x})$ for the latent variables to approximately update the intractable likelihood term \cite{kingma2013auto,rezende2014stochastic,kim2018tutorial}, whose architecture based on neural network methods and variational inference is in Figure~\ref{fig:workscheme}.
\subsubsection{Variational Auto-Encoder (VAE) \cite{bowman2015generating}}
Variational auto-encoder (VAE) explores the hidden patterns of texts by adding a latent code regularization term to a plain auto-encoder. Concretely speaking, the latent variable $\mathbf{z}$ is contributed by optimizing the evidence lower bound (ELBO), which takes both reconstruction error and a regulation loss implemented by Kullback-Leibler divergence (KLD) into count:
\begin{equation}
\begin{aligned}
\log p_{\theta}(\mathbf{X}) & \geq \underbrace{\mathbb{E}_{q_{\phi}(\mathbf{z} \mid \mathbf{x})}\left[\log p_{\theta}(\mathbf{x} \mid \mathbf{z}\right)] }_{\text{\small{reconstruction term}}}\\& -\underbrace{\mathbb{D}_{\mathrm{KL}}\left(q_{\phi}(\mathbf{z} \mid \mathbf{x}) \| p_{\theta}(\mathbf{z})\right)}_{\text{\small{regularization term}}},
\end{aligned}
\label{eq: E}
\end{equation}
where network parameter $\phi, \theta$ belongs to the encoder and decoder respectively to produce latent posterior and prior.
Despite its success in depicting the complex distribution of data and sampling from it to generate diverse and fluent content, they have apparent defects when it comes to conditional language modeling without extra side information:
\begin{enumerate}
    \item KL collapse problem \cite{bowman2015generating}: caused by an overmighty decoder of VAE and leads the model to ignore learned knowledge from latent space.
    \item Token-latent irrelevant issue \cite{shen2020educating}: sentences that are similar to each other may share no connection in their latent representations.
    \item Latent vacancy issue \cite{xu2020variational}: latent variables with knowledge for disentangling text features spread discretely in the hidden space.
\end{enumerate}
Several approaches have been devised to handle the first issue, including optimizing decoder architectures \cite{yang2017improved,semeniuta2017hybrid}, inventing auxiliary objectives \cite{zhao2017infovae,zhao2017learning,xiao2018dirichlet}, novel encoder training schedule \cite{bowman2015generating,fu2019cyclical}, flexible posterior~\cite{tomczak2016improving,wang2019topic}, etc. These methods generally share the same goal: to impair the ability of a powerful recurrent decoder and enhance the utterance of latent space. The former mitigates this puzzle mainly by weakening the conditional dependency of VAE on its decoder, but it may fail to generate high-quality continuous sentences. And methods from the latter one strengthen latent expression, and make the hidden states compatible with the powerful decoder in VAE. Except for KL collapse problem, the last two remained issues are the main reasons to explain why the vanilla text VAE cannot be directly interpolated to be controllable like several parallel applications in image field \cite{higgins2016beta,chen2018isolating}. 
\subsubsection{Adversarial Auto-Encoder (AAE) \cite{makhzani2015adversarial}}
Adversarial auto-encoder and Wasserstein auto-encoder (WAE) \cite{tolstikhin2017wasserstein} generally share a common goal with VAE in employing the ELBO maximization to update the holistic model. A major distinction between these two kinds of models lies in the regularization term of their ELBOs. While VAE takes a Kullback-Leibler (KL) penalty as its latent regulator, AAE (or WAE) introduces a discriminator to judge latent differences as illustrated below: 
\begin{equation}
    \begin{aligned}
    &\log P_{\theta}(\mathbf{x}) \geq \underbrace{\mathbb{E}_{q_{\phi}(\mathbf{z} \mid \mathbf{x})}\left[\log p_{\theta}(\mathbf{x} \mid \mathbf{z}\right)] }_{\text{\small{reconstruction term}}}
    \\&-\underbrace{\mathbb{E}_{p_{\theta_1}(\mathbf{z})}[-\log D(\mathbf{z})]+\mathbb{E}_{p_{\theta_2}(\mathbf{x})}[-\log (1-D(E(\mathbf{x})))]}_{\text{\small{discriminator penalty}}},
    \end{aligned}
    \label{eq0}
\end{equation}
where function $D(\cdot)$ and $E(\cdot)$ for the ELBO in AAE (or WAE) denote its discriminator and encoder respectively, and $\theta_1, \theta_2$ denote parameters of model decoder and discriminator, model decoder, discriminator and encoder accordingly. In contrast to VAEs, AAEs (or WAEs) maintain a strong coupling between their hidden codes and decoder by applying theoretically superior distance measurement (e.g., Wasserstein distance), ensuring that the decoder does not ignore representations in the latent space, which makes it robust for latent knowledge interpretation and interpolation \cite{vincent2010stacked,makhzani2015adversarial}.

\subsubsection{Discrete Variational Auto-Encoder \cite{oord2017neural}}
Turning the continuous latent space into a discrete one is favorable in indicating latent knowledge regardless of the strong decoder of VAE and avoiding the KL collapse problem for good. Vector Quantized Auto-Encoder (VQ-VAE) was therefore invented. It has the same training paradigm as VAE and turns continuous hidden variables into discrete vectors by looking up a discrete codebook $\mathbf{c}\in \mathbb{R}^{K\times D}$ that obtains in advance, where $K$ and $D$ represents the size of the discrete latent space and dimensionality of each embedding vector $\mathbf{c_i}$ severally. The posterior categorical distribution $q(\mathbf{z}\mid \mathbf{x})$ is then defined as:
\begin{equation}
    q(\mathbf{z} = k \mid \mathbf{x})=\left\{\begin{array}{ll}1 & \text{for } k=\operatorname{argmin}_{j}\left\|E(\mathbf{x})-\mathbf{c_{j}}\right\|_{2} \\ 0 & \text{ otherwise }\end{array}\right.,
\label{eq:vq-vae}
\end{equation}
here $E(\mathbf{x})$ is the output of the encoder of VAE. After the vector quantification in Eq. (\ref{eq:vq-vae}), every latent code in VQ-VAE is in discrete mode through one-hot projection. Since the latent quantization process cannot derive gradients because the $\operatorname{argmin}$ operation exists, VQ-VAE employs straight-through estimator \cite{bengio2013estimating} to approximate the gradient and just copy gradients from decoder input to encoder output $E(\mathbf{x})$. 

\subsubsection{Variational Auto-Encoder with Pre-trained Language Model}
Large Pre-trained language models (PLMs) are gaining more and more popularity these days. With enormous resources being devoted, well-experienced encoders/decoders such as BERT \cite{devlin2018bert} and GPT-2 \cite{radford2019language} are devised to fully understand textual content and create human-like sentences respectively. Incorporating such mighty PLMs as both encoder and decoder of a variational auto-encoder can largely mitigate the KL collapse problem by offering the decoder a nonnegligible latent space from its encoder \cite{li2020optimus}. How to take full advantage of these PLMs to variational auto-encoders has been explored nowadays \cite{liu2019transformer,li2020optimus,fang2021transformer,park2021finetuning,tu2022adavae}, which have shown promising potential in a varied multitude of tasks including unsupervised latent interpolation and semi-supervised conditional story generation, etc. Since PLMs were generally pre-trained in large corpus without annotations before applying to specific downstream tasks, the training mode of models with PLMs is essentially semi-supervised if the models' fine-tuning stage requires explicit labels. We categorize them as semi-supervised in Section \ref{semi}.

\section{Methodologies}
In this section, we will introduce specific methods that aim at controllable text generation using auto-encoders. We divide this section according to the training paradigm of existing models, namely supervised, semi-supervised and unsupervised. As for concrete experimental setups, supervised controllable models require their training data strictly obey the form of (text, label) pairs. Semi-supervised methods only need partial training data to follow such form and the rest can be textual contexts without annotations. The unsupervised method is the extreme case that asks for no annotation of the training corpus. A full list of methods is presented below: 

\begin{table*}[ht]
\begin{tabular}{l|l}
\toprule[1.5pt]
Training Paradigm        & Methodologies                                                                                                                                                                                                                                                                                                                                                                                   \\ \midrule
Supervised               & \begin{tabular}[c]{@{}l@{}}\citet{wang2019controllable}, \citet{zhang2019improve},\\ \citet{shao2019long}, \citet{fang-etal-2022-controlled} \end{tabular}                                                                                                                                                                                                                                                                 \\ \midrule
Semi-Supervised          & \begin{tabular}[c]{@{}l@{}}\citet{yang2017improved}, \citet{logeswaran2018content},\\ \citet{zhao2018adversarially}, \citet{subramanian2018towards}, \citet{li2020optimus}, \\\citet{ye2020variational}, \citet{mai2020plug}, \citet{cheng2020improving}, \\\citet{duan2020pre}, \citet{mai2021bag}, \citet{fang2021transformer} \\ \citet{tu2022pcae} \end{tabular} \\ \midrule
Unsupervised (w/ \textit{topic})  & \begin{tabular}[c]{@{}l@{}}\citet{xiao2018dirichlet}, \citet{wang2018topic},\\ \citet{bao2019generating},  \citet{wang2019topic}, \citet{tang2019topic},\\ \citet{xu2020variational}\end{tabular}                                                                                           \\
Unsupervised (w/o \textit{topic}) & \begin{tabular}[c]{@{}l@{}}\citet{ghabussi2019stylized}, \citet{fang2019implicit},\\ \citet{shi2020dispersed}, \citet{shen2020educating}, \citet{rezaee2020discrete}\\ \citet{li2020optimus}, \citet{mercatali2021disentangling} \citet{tu2022adavae} \\ \citet{hu2022fuse} \end{tabular}                                                                    \\
\bottomrule[1.5pt]
\end{tabular}
\label{tab0}
\caption{Methodologies of controllable text generation via variational auto-encoders. Classified by training manners. We present \textit{topic} attribute for models with or without explicit topic modeling part, see Section \ref{unsup} for details.}
\end{table*}

\subsection{Supervised Methodologies}\label{supmethods}
The problem is how to make latent AEs, the amazing generators, to be controllable. The most intuitive and straightforward solution is to merge all labels (or conditions) into its generation process. Conditional VAE (CVAE) \cite{sohn2015learning} was proposed following this thought. For the given condition $\mathbf{c} = {c_1, c_2, ..., c_n}$ with $n$ to be the sample size. By adding the condition to both encoder and decoder of VAE, the overall generation process is compelled to take $\mathbf{z}$ and $\mathbf{c}$ into consideration, that is to say, the posterior and prior of latent distribution are crafted to be compatible with given conditions. This can be reflected in the ELBO of CVAE:
\begin{equation}
    \begin{aligned}
    \log P_{\theta}(\mathbf{x}\mid \mathbf{c}) & \geq \mathbb{E}_{q_{\phi}(\mathbf{z} \mid \mathbf{x}, \mathbf{c})}\left[\log p_{\theta}(\mathbf{x} \mid \mathbf{z},\mathbf{c}\right)]\\&-\mathbb{D}_{\mathrm{KL}}\left(q_{\phi}(\mathbf{z} \mid \mathbf{x},\mathbf{c}) \| p_{\theta}(\mathbf{z}\mid \mathbf{c})\right).
\end{aligned}
\end{equation}
One issue that becomes serious in CVAE is the KL collapse problem. For a vanilla VAE, it requires its latent $\mathbf{z}$ to obey the prior distribution $p(\mathbf{z})$, which causes its decoder to ignore the encoder's explanation of data. CVAE merges condition straight to the decoding process and unwittingly encourages the decoder to ignore the information in latent space and eventually triggers KL collapse. As a fix-up, \citet{zhang2019improve} proposed to construct ``perfect'' encoders concerning different decoders in order to infuse $\mathbf{x}$ into latent $\mathbf{z}$ to evade the standing of $q(\mathbf{z}\mid x) = p(\mathbf{z})$. Formally, \citet{zhang2019improve} introduced a self-labeling component with $\mathbf{x}$ and condition $\mathbf{c}$ as input to approximate the reverse process $g(\mathbf{x}, \mathbf{c})$ of decoder and produce self-labeled latent codes $\mathbf{z_{\text{label}}} := g(\mathbf{x}, \mathbf{c})$. By taming latent code $\mathbf{z}$ to conclude input knowledge, the objective of such CVAE becomes:
\begin{equation}
    \begin{aligned}
    \log P_{\theta}(\mathbf{x}\mid \mathbf{c}) &\geq \mathbb{E}_{q_{\phi}(\mathbf{z} \mid \mathbf{x}, \mathbf{c})}\left[\log p_{\theta}(\mathbf{x} \mid \mathbf{z},\mathbf{c}\right)]\\& -\mathbb{D}_{\mathrm{KL}}\left(q_{\phi}(\mathbf{z} \mid \mathbf{x},\mathbf{c}) \| p_{\theta}(\mathbf{z}\mid \mathbf{c})\right)\\&- \lambda \mathbb{E}_{q_{\phi}(\mathbf{z} \mid \mathbf{x}, \mathbf{c})}\|\mathbf{z} - \mathbf{z_{\text{label}}}\|^2,
    \end{aligned}
\end{equation}
which yields generation ability with high diversity, demonstrates that this CVAE does not fall into the pitfall of the monotonous latent posterior distribution. Though CVAE can be controlled by incorporating condition knowledge, its training scheme inevitably requires full annotation with one label per document and confines CVAE into the generation with global soft constraints. Approaches with full supervision can perform much better in either content quality or control proficiency. 

\begin{figure*}[ht]
\centering
\includegraphics[width=1.\linewidth]{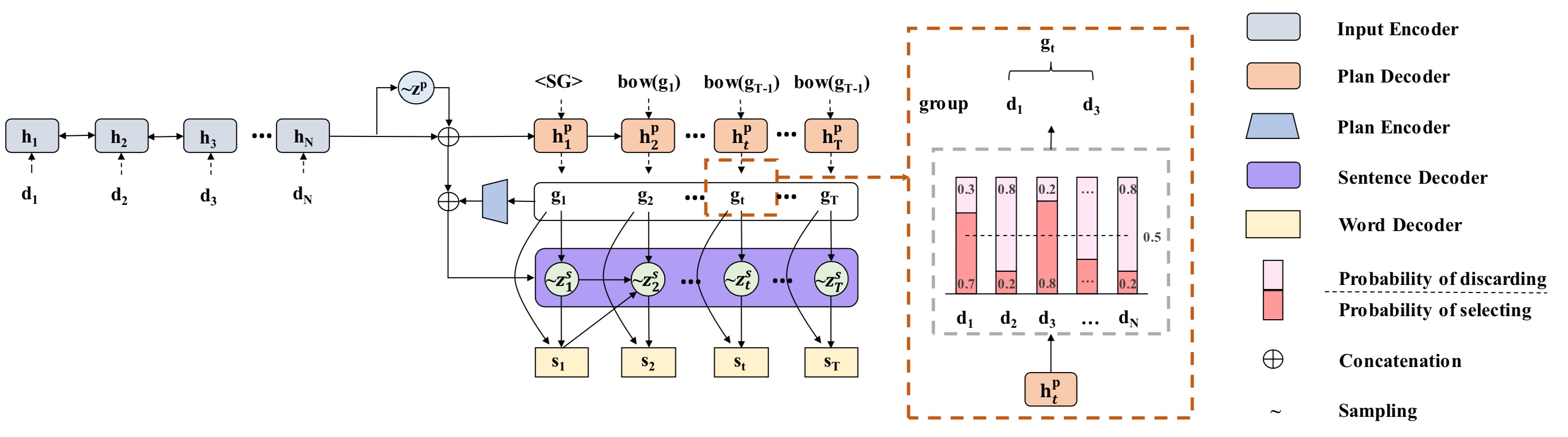}
\caption{Model framework of PHVM from \citet{shao2019long}. It leverages two LVMs and two latent spaces to guide planning and sentence generation respectively. $\mathbf{z^p}$ is the latent code for global planning, while $\mathbf{z_t^s}$ is the latent code for sentence-level generation at timestep $t$.}
\label{fig:phvm}
\end{figure*}

To a more fine-grained controllable generation, \citet{shao2019long} first proposed the PHVM model to employ the latent variable model in table-to-text generation with hard control. Specifically, as shown in Figure \ref{fig:phvm}, PHVM adopts two VAEs for keyword planning and sentence generation respectively, which gives birth to two separate types of latent codes: planning-level $\mathbf{z^p}$ and sentence-level $\mathbf{z^s}$. While $\mathbf{z^p}$ is responsible for producing various plans for sentences with given keywords, $\mathbf{z^s}$ is made for sentence generation with pre-assigned keyword planning (regarded as a CVAE with fine-grained labels). This setting derives two ELBOs with regard to $\mathbf{z^p}$ and $\mathbf{z^s}$ severally. Moreover, PHVM uses bag-of-word loss \cite{zhao2017learning} to alleviate KL collapse and employs hierarchical generation (i.e., models both sentence-level and word level sequential information), which makes it competent to produce diverse and long texts with hard control of keywords. PHVM provides us with a new direction for applying variational auto-encoders, which is not constricted in generating sentences but flexible keyword plans. 

Inspired by discrete VAEs such as VQ-VAE \cite{oord2017neural}, \citet{fang-etal-2022-controlled} recently propose a framework with discrete latent prior weighted by continuous Dirichlet distribution named Dprior. With every given label represented by a discrete latent variable, the controllable generation process can be easily accomplished by choosing the exact subset of the latent variable. To back-propagate the discrete latent variables during training, Dprior develops the dual function of KL divergence based on iVAE \cite{fang2019implicit}. \citet{fang-etal-2022-controlled} further extend the model in both LSTM and PLM domains with supervised contrastive objective \cite{van2018representation} to enhance the model's controllability. \citet{fang-etal-2022-controlled} presents a new stage of developing VAE models aim for controllable text generation: one is to verify the model with both LSTM and PLM encoder\&decoder separately, the other one is to incorporate the advanced learning tricks such as auxiliary losses to enhance model performance.

However, these methods can only be applied on datasets with the full-size labels, which becomes unprofitable in real-world circumstances where massive accurate annotations are expensive and rare.
\subsection{Semi-supervised Methodologies} \label{semi}
Controllable variational auto-encoders under the semi-supervised paradigm only require partial annotations of the training data. Semi-VAE (SVAE) was proposed by \citet{kingma2014semi} and applied in the vision domain. Recently, it was extended to controllable text generation as a baseline \cite{duan2020pre}. For labeled data, SVAE takes $\mathbf{z}$ for content modeling, $\mathbf{y}$ as label embedding, and follows the training path of CVAE. For unlabeled data, SVAE treats label latent embedding $\mathbf{y}$ as $\mathbf{z}$ and updates them equally. SVAE can also take a pre-trained plain VAE and incorporates the wake-sleep algorithm \cite{dayan2000helmholtz} to acquire reliable latent representations and produces authentic content. \citet{yang2017improved} adopted a similar training structure with SVAE (treated the label embedding as latent codes), but a dilated convolutional neural network \cite{yu2015multi} with residual connection as the decoder. This modification theoretically weakens the ability of VAE's decoder and averts the KL vanishing problem as expected. To improve the convenience of deriving meaningful latent knowledge on different domains, \citet{zhao2018adversarially} further generalized the two-stage training framework for controllable latent vector production. They used unlabeled contexts for auto-encoder pre-training and devised a conditional GAN on latent space in order to obtain latent vectors to generate controllable texts. Their method made progress in both visual and textual generation tasks. Despite their successes, the wake-sleep algorithm and two-stage training inevitably require two round of full training of VAEs. This is not portable and may consume a huge amount of resources in this process.
\begin{figure*}[ht]
\centering
\includegraphics[width=1.\linewidth]{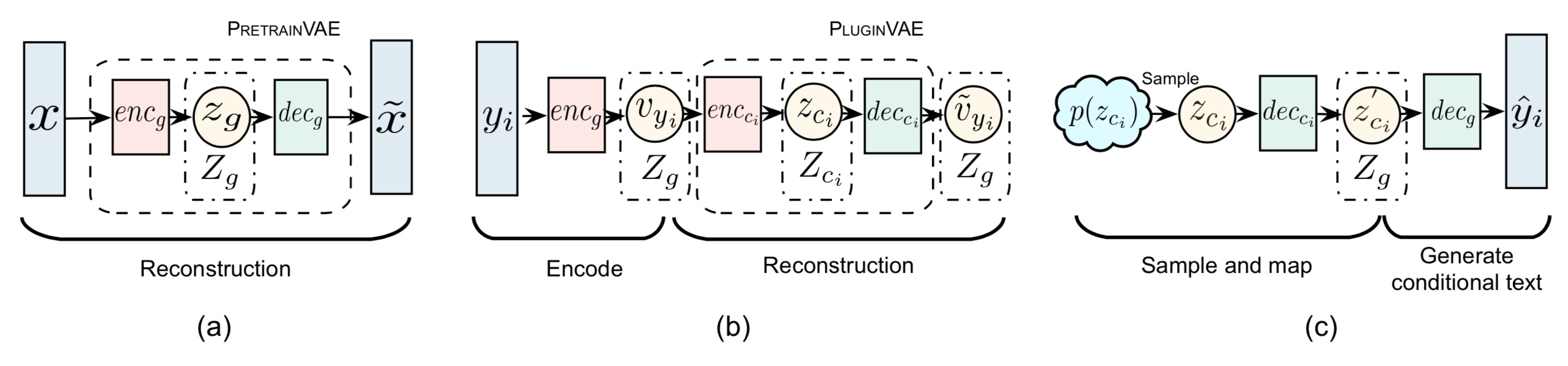}
\caption{Model framework of PPVAE from \citet{duan2020pre}. Its training/generation paradigm consists of three parts (a) pre-training \texttt{PretrainVAE} on the large unlabeled corpus to obtain $\mathcal{Z}_g$; (b) Train the \texttt{PluginVAE} with only latent space projections being activated on the labeled corpus to obtain $\mathcal{Z}_c$; (c) Sample the latent code from conditional before generate controllable texts.}
\label{fig:ppvae}
\end{figure*}
\citet{duan2020pre} proposed a novel framework for text generation named ``pre-train and plug-in'' (also known as ``plug-and-play'' or PnP for short) in Figure \ref{fig:ppvae}. Their model PPVAE is consists of two separate parts: a pre-trained WAE with global latent space $\mathcal{Z}_g$ and a plug-in VAE with constrained latent space $\mathcal{Z}_c$. While the pre-trained WAE is trained in advance without any restriction, plug-in VAE is actually an extension of pre-trained WAE with very few added trainable parameters whose goal is to construct a sensible mapping from $\mathcal{Z}_g$ to $\mathcal{Z}_c$ under the paradigm of VAE. During the training of plug-in VAE, all parameters in the pre-trained model are frozen, which makes it super fast to converge with very few biased samples (samples that belong to the same category) as input. As an extension, \citet{mai2020plug} invented \textit{emb2emb} model for text style transfer generation under the PnP framework, which devised a novel ``offset net'' as a mapping function that only deploys few arguments in the latent manifold. By introducing \textit{Broadcast Net} to both RNN-based and pre-trained auto-encoders, \citet{tu2022pcae} further lead the PnP controllable auto-encoders to a more flexible and efficient direction with their PCAE framework. Its vital \textit{Broadcast Net} creates a compact and manipulable controllable latent space by repeatedly adding label signals to the original latent manifold during plug-in process, reaching a higher degree of generation control and fewer training costs compared with previous work.

To theoretically explain the controllable signal flow in latent AEs, \citet{cheng2020improving} resorts to information theory for disentangled representation learning (DRL) in controllable VAE. They began by estimating the variation of information between target sentence $\mathbf{x}$ and its style label $y$, split latent code into $\mathbf{c}$ for content embedding and $\mathbf{s}$ for style embedding respectively, then derived a disentangling learning loss in addition to the original ELBO:
\begin{equation}
    \begin{aligned}
    \mathcal{L}_{\text{Dis}}:=\mathrm{I}(\mathbf{s} ; \mathbf{c})&-\mathbb{E}_{p(\mathbf{x}, \mathbf{c})}\left[\log q_{\phi}(\mathbf{x} \mid \mathbf{c})\right] \\&-\mathbb{E}_{p(y, \mathbf{s})}\left[\log q_{\psi}(y \mid \mathbf{s})\right],
    \end{aligned}
\end{equation}
where the distribution $p(\mathbf{x}\mid \mathbf{c}, p(y\mid \mathbf{s})$ are parameterized by a recurrent decoder and a classifier severally. Besides, they brought another decoder into consideration to fulfill the reconstruction term related to input sequences that involves both $\mathbf{s}, \mathbf{c}$ in their model.

These semi-supervised methods generally bound to produce softly controllable texts with global labels. As the fine-grained controllable generation with hard constraints, \citet{ye2020variational} proposed a novel framework VTM to employ two VAEs for both table template and textual content modeling. That is to say, for a given table-text pair $(\mathbf{x}, \mathbf{y})$, VTM requires a latent $\mathbf{z}$ for the template and another latent $\mathbf{c}$ for content. its objective is described as:
\begin{equation}
    \begin{aligned}
    & \log P_{\theta}(\mathbf{y}\mid \mathbf{x}) \\&\geq \mathbb{E}_{q_{\phi}(\mathbf{z} \mid \mathbf{y})}\left[\log p_{\theta}(\mathbf{y} \mid \mathbf{z},\mathbf{c} = f_{\text{enc}}(\mathbf{x}),\mathbf{x}\right)]\\& -\mathbb{D}_{\mathrm{KL}}\left(q_{\phi}(\mathbf{z} \mid \mathbf{y}) \| p_{\theta}(\mathbf{z})\right)\\&
    -\mathbb{D}_{\mathrm{KL}}\left(q_{\phi}(\mathbf{c} \mid \mathbf{y}) \| p_{\theta}(\mathbf{c})\right),
    \end{aligned}
\end{equation}
where $f_{\text{enc}}(\cdot)$ is the encoder function for table embedding and the prior $p_{\theta}(\mathbf{c}) = \delta(\mathbf{c} = f_{\text{enc}}(\mathbf{x}))$ ensures the assumption of exact match of table-text pair. In addition, they appended several auxiliary losses out of content and template preserving purposes. For raw text without table description, latent representation $\mathbf{c}$ is sampled from a normal Gaussian $N(0, I)$ as a generalization.

Most recently, with the rise of large pre-trained language models (PLMs), more and more work tends to incline promising encoder/decoder such as GPT-2 \cite{radford2019language} to create realistic sentences. Controllable language models built on PLMs often fine-tune PLMs as part of the model structure and feed labels in this process for immediate applications. Since PLMs are usually trained with massive unlabeled corpus, the whole controllable model only accesses to label information at fine-tuning stage, thus making it semi-supervised. \citet{li2020optimus} was the first one to connect pre-trained BERT and GPT-2 via continuous latent space following the VAE paradigm. They achieved state-of-the-art controllability in text generation via a two-stage training described in \citet{zhao2018adversarially}, which is largely ascribed to its competent encoder and decoder components. \citet{fang2021transformer} proposed the first CVAE based on PLMs. Their proposed VAE architecture is modified by replacing the original encoder and decoder to pre-trained GPT-2s. Unlike a plain VAE that easily mixes latent codes into the decoder input. \citet{fang2021transformer} explored several combination rules for infusing latent codes into the framework of Transformer-based AE, including at the beginning of word embedding layer, in the middle of attention hidden states, and at the last word decoding process. As for controllable generation, they fine-tuned the holistic model with given labeled prompts to produce conditional texts. It is no doubt that these new methods gave us a hint about the application to employ PLMs into variational auto-encoders, they were short of detailed discussions about the condition incorporation methods into such a large language model to make them learn obedience.

\subsection{Unsupervised Methodologies}\label{unsup}
While supervised signals assist models to put out content with accurate attributes, an unsupervised manner is the way a plain auto-encoder adopts. Once we obtain reliable latent information for disentangled representations via unsupervised training, it is facile to take full advantage of them under any circumstance and benefits several downstream tasks (text summarization \cite{wang2019topic}, style transfer \cite{hu2017toward,tang2019topic},  translation task \cite{liu2018learning}, etc.) effectively. Intuitively, the target of producing topic-specified sentences can fall into three courses: topic extraction, sequential learning, and joint generation. Therefore, both topic and sequential models are of great importance in analyzing and creating controllable texts. In general, the ways to learn and comprise topic information in the text generation system can be divided into two schemes: models with explicitly additional topic modeling parts and models without them.
\subsubsection{$\textbf{w/}$ Explicit Topic Modeling Part}
Explicit topic modeling parts are designed to separate topic learning and sequential learning process. The input of this part generally follows Bag-of-Word (BoW) manner to cancel influences brought by the sequential connection between words. And the model structure of the topic modeling part varies from the plain LDA to well-designed neural networks. D-VAE \cite{xiao2018dirichlet} firstly proposed to model unsupervised latent model with the topic latent model (specifically an LDA model) conditioned on latent code $\mathbf{z}$. Formally, \citet{xiao2018dirichlet} investigated two modes for topic information infusion: a two-stage model with Dirichlet latent from a pre-trained LDA and an end-to-end model with topic latent code $\mathbf{t}\sim \text{Dirichlet}$ and being conditioned on $\mathbf{z}$ to acquire text knowledge upfront. The holistic ELBO is then derived as:
\begin{equation}
    \begin{aligned} 
    \log p(\mathbf{x}) \geq & \mathbb{E}_{q(\mathbf{z}, \mathbf{t} \mid \mathbf{x})}[\log p(\mathbf{x} \mid \mathbf{z}, \mathbf{t})]\\&-\mathbb{D}_{\mathrm{KL}}(q(\mathbf{z}, \mathbf{t} \mid \mathbf{x}) \| p(\mathbf{z}, \mathbf{t})) \\=& \mathbb{E}_{q(\mathbf{z} \mid \mathbf{x})}\left[\mathbb{E}_{q(\mathbf{t} \mid \mathbf{x}, \mathbf{z})}[\log p(\mathbf{x} \mid \mathbf{z}, \mathbf{t})]\right]\\&-\mathbb{E}_{q(\mathbf{z} \mid \mathbf{x})}\left[\mathrm{D}_{\mathrm{KL}}(q(\mathbf{t} \mid \mathbf{x}, \mathbf{z}) \| p(\mathbf{t} \mid \mathbf{z}))\right] \\&- \mathbb{D}_{\mathrm{KL}}(q(\mathbf{z} \mid \mathbf{x}) \| p(\mathbf{z})). \end{aligned}
\end{equation}
At the same time, \citet{wang2018topic} focused on neural topic modeling and invented a Gaussian-based neural topic component with a mixture of expert (MoE) \cite{hu1997patient} decoder (TCNLM), so each expert concentrates on one dimension of the latent topic variable and produces controllable sentences. To make full use of the nature of variational auto-encoder in accurately characterizing data patterns, \citet{wang2019topic} went to propose TGVAE based on the exact neural topic model in TCNLM. The major improvement of TGVAE compared with TCNLM lies in its latent design. While TCNLM employs a mixture of agents on the token generation level (MoE decoder), TGVAE leans upon a mixture of agents on the latent level, that is Gaussian mixture model (GMM) for latent modeling. In detail, the prior of latent GMM in TGVAE is parameterized by the latent variable from its neural topic component, and it employs Householder flow \cite{tomczak2016improving} for flexible GMM posterior estimation, a novel KL divergence to calculate the latent regularization term. With each Gaussian representing a topic in the data, it produces controllable contexts by manipulating distinct Gaussian distributions in the latent space.
\begin{figure*}[ht]
\centering
\includegraphics[width=0.9\linewidth]{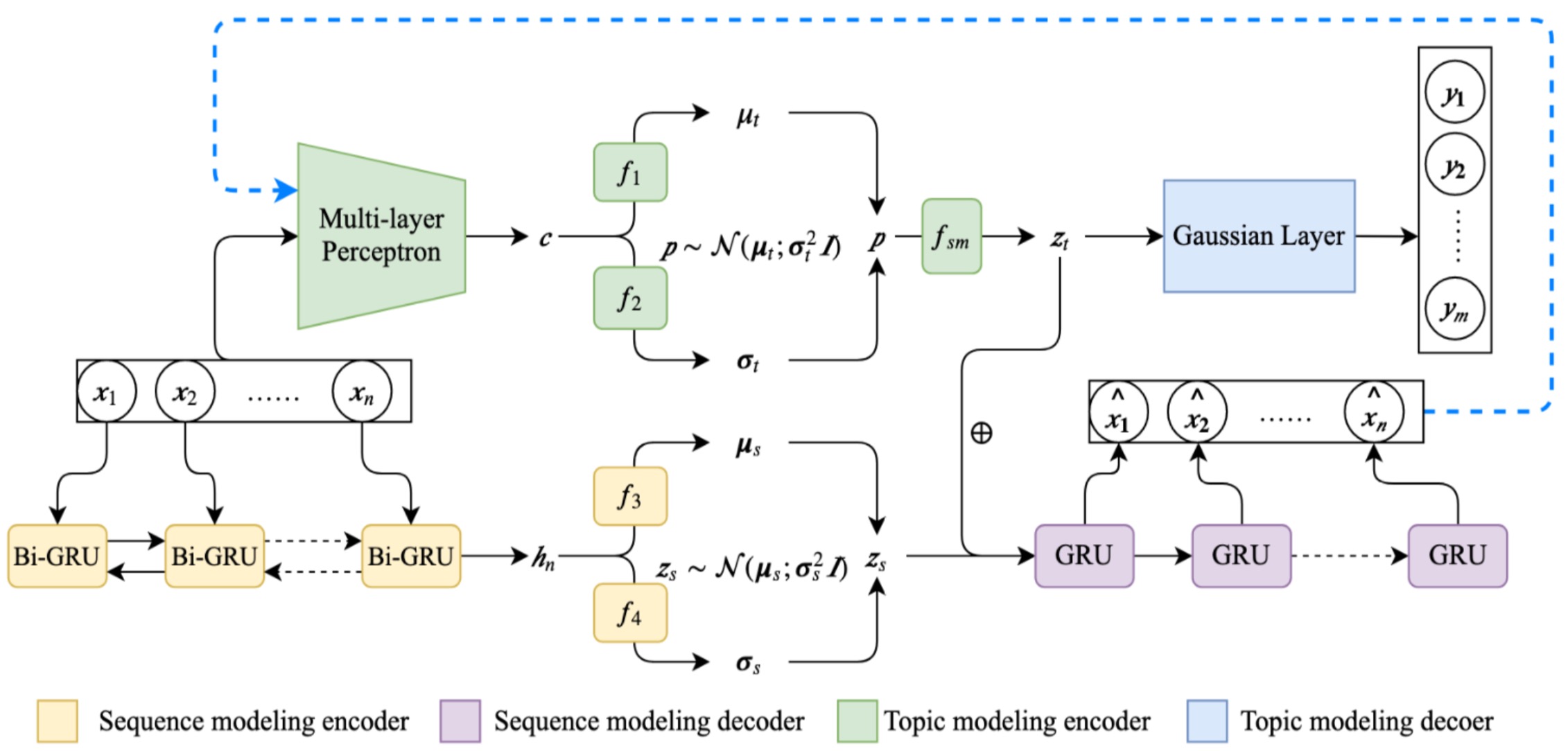}
\caption{Model framework of TATGM from \citet{tang2019topic}. TATGM consists of an explicit topic modeling part and the sequence modeling part. Both parts are trained jointly via two VAEs, and both $\mathbf{z_t}$, $\mathbf{z_s}$ follow the isotropic Gaussians. The topic modeling part also plays the role of a topic word discriminator.}
\label{fig:tatgm}
\end{figure*}
These two methods confound the topic and content latent field at the beginning of the text generation process, thus being less explainable compared with D-VAE. Other approaches that manage content and topic separately arose, \citet{tang2019topic} proposed TATGM with $\mathbf{z_t}$ and $\mathbf{z_s}$ that both follow an isotropic Gaussian to the model topic and sequence representations severally in Figure \ref{fig:tatgm}. They fed sentences in the Bag-of-Word (BoW) manner into the topic encoder and made it a discriminator for topic modeling augmentation. By concatenating $\mathbf{z_s}$ and $\mathbf{z_t}$, TATGM is capable of producing controllable textual content. A concurrent work from DSS-VAE \cite{bao2019generating} described the input sentences of sequence latent space in the schema of linearized tree sequence \cite{vinyals2015grammar}. Notably, DSS-VAE does not use BoW as a special input way for topic modeling, but adopts several auxiliary adversarial losses to ensure responsible stages of disentangled learning and combined generation between $\mathbf{z_t}$ and $\mathbf{z_s}$. These methods join additional topic modeling parts into generation following the intuition that text semantic and syntax structures are inherently distinct.

In the field of computer vision, $\beta$-VAE \cite{higgins2016beta} and its derivations are famous for only interpolating the latent space in a vanilla VAE to produce attribute-specified images. However, \citet{xu2020variational} proved the infeasibility to conduct similar experiments with text VAE, they presented that variational auto-encoders that perform in a discrete domain (e.g., sentence) are unable to be generalized likewise because of the latent vacancy problem in their latent spaces. They split the latent space into two parts $\mathbf{z^{(1)}}, \mathbf{z^{(2)}}$ with word embedding and RNN embedding as initial state respectively, and borrowed principles in simplex theory to constrain $\mathbf{z^{(1)}}$ in a simplex manifold for topic knowledge induction as well as additionally derived losses.
\subsubsection{$\textbf{w/o}$ Explicit Topic Modeling Part}
Without an explicit topic modeling part, a VAE model is required to learn both topic and sequential knowledge with one confined latent space. This demand rises a big challenge for the vanilla text VAE that has several latent limitations. \citet{xu2020variational} identified the possible reason for vanilla variational auto-encoder not being able to control in the fashion of $\beta$-VAE, they still designed topic information learning specially and imposed restrictions only on topic latent code (e.g., constraints on $\mathbf{z^{(1)}}$). Introducing an accessional component for explicit topic depiction is one way to make generated text controlled. The question is, are there more general methods that do not compulsively require extra components based on variational auto-encoders? 

For optimization on the overall latent level, \citet{ghabussi2019stylized} proposed to use GMM in the latent space as a replacement of isotropic Gaussian, biased samples were used for training model with corresponding Gaussian distribution activated (i.e., mixture weight was not set to zero). When it comes to controllable inference, this model reaches such a goal by sampling from chosen Gaussian distributions. Besides, they also employed WAE in order to conduct robust topic information. As a way to improve, \citet{shi2020dispersed} explored several different distribution types (e.g., exponential family) with a mixture model and aimed at learning structured topic pattern without any biased data. They identified the ``mode collapsed'' problem related to a mixture of distributions in the text variational auto-encoders and further devised regularization terms as a countermeasure.

\begin{figure*}[ht]
\centering
\includegraphics[width=1.0\linewidth]{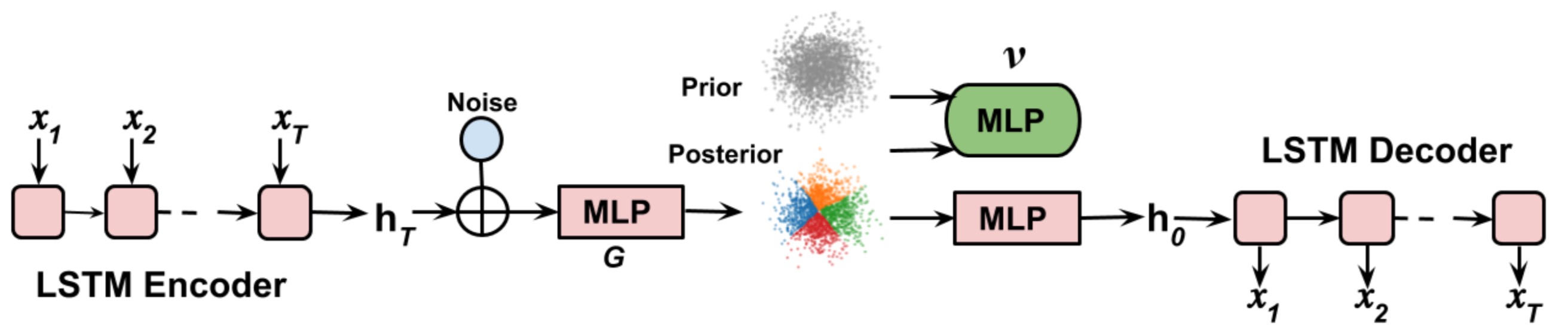}
\caption{Model framework of i-VAE from \citet{fang2019implicit}. i-VAE adds the random noise before the reparamterization of the latent space. It also derives a mutual information term between the input and the latent code. These modifications make the latent representation in i-VAE more robust and smooth to interpolate.}
\label{fig:ivae}
\end{figure*}

For a more generalized extension, \citet{fang2019implicit} proposed an innovative informative penalty $I(\mathbf{x}; \mathbf{z})$ as well as a noise-adding procedure that can be applied to any LVM to close the gap between observed data $\mathbf{x}$ and latent code $\mathbf{z}$ as in Figure \ref{fig:ivae}. Their original intention is similar to the one in \citet{zhang2019improve} as introduced in Section \ref{supmethods}, but used a more intuitive yet effective approach. \citet{shen2020educating} provided another view to construct the hidden space of text variational auto-encoders that allow us to interpolate to produce controllable sentences. They found that the semantic similarity of sentences has almost nothing to do with their similarity score mapped in the latent field, simply adding noise (e.g., word dropout, word substitution) at the input token level can effectively alleviate such conundrum. By doing so, there stands a chance to control textual output by means of manipulating the refined latent space. Besides adding adversarial noises to the latent space to help text VAEs to be controllable without additional modifications on the model level. VAEs with strong (pre-trained) encoders that are compatible with their decoder modules can also mitigate challenges faced by a plain text VAE. Optimus proposed by \citet{li2020optimus} also displays its competence for unsupervised controllable text generation by simply manipulating latent space. Further, \citet{tu2022adavae} came up with a more efficient big VAE model AdaVAE with GPT-2 as its encoder and decoder. They only activate the adapter \cite{houlsby2019parameter} components inside the VAE during training, accomplishing sensibly satisfactory controllable results in latent manipulation and interpolation tasks. Another direction to produce controllable texts without supervision lies in fusing learned latent knowledge into the decoder with more confidence. \citet{hu2022fuse} proposed a novel knowledge infusion method to inject latent information into its transformer decoder layer by layer, resulting in robust text reconstruction performance. This ability helps transformer-based VAE models produce consistent sentences when manipulating their latent spaces.

In order to help the latent space be not only controllable but more explicable, \citet{rezaee2020discrete} proposed VRTM, which turned the conventional continuous latent space into a discrete one. They allocated a latent code for each word and predict its topic polarity (i.e., binary sign) through additional Dirichlet variables inferred from a masked BoW embedding (mask non-thematic words). These settings give them permission to measure topic distributions on both word level and document level, so they could naturally control textual output with more flexibility. Most recently, another work conducts on discrete text VAE appears. \citet{mercatali2021disentangling} focused on the decomposition of a VAE's ELBO, which mainly followed work that were previously carried out in the computer vision field \cite{chen2018isolating}. Extending this work to a discrete text VAE, \citet{mercatali2021disentangling} helps us to understand responsible disentangling factors in the variational auto-encoders, and also to identify their own roles in the controllable text generation process.

\section{Metrics}
Evaluation metrics are roughly divided into two parts: language generation metrics for evaluating models' generative capacity in the language domain and control ability metrics for testing the control degree of models.
\subsection{Language Generation Metrics}
Controllable text generation as a language generation task, its evaluation standards are quite similar to language modeling via latent variable models.
\begin{table*}[!t]
\centering
\renewcommand\arraystretch{1.0}
\begin{tabular}{c|cccccc}
\toprule[1.5pt]
\textbf{Model}                                & \textbf{APNEWS}                 & \textbf{IMDB}                   & \textbf{BNC}                    & \textbf{PTB}   & \textbf{Yahoo}     & \textbf{Yelp15}     \\ \midrule
\textbf{LSTM LM}                             &64.13                            &72.14                
      &102.89                         &116.2          &66.2  &42.6          \\
\textbf{LSTM+LDA}                             &57.05                           &69.58                
       &96.42                         &-        &53.5 &37.2           \\
\textbf{LSTM VAE} \cite{bowman2015generating}                             &75.89                            &86.16                
      &105.10                        &96.0        &65.6 &45.5              \\
\textbf{VAE+HF}                             &71.60                            &83.67                
       &104.82                        & -        & - & -              \\
\textbf{DCNN-VAE} \cite{yang2017improved}                                 & -                               & -                               & -                               & -        & 63.9 & 41.1            \\
\textbf{TCNLM} \cite{wang2018topic}                                & 52.75                           & 63.98                           & 87.98                           & -            & - & -           \\
\textbf{DVAE} \cite{xiao2018dirichlet}                                 & -                               & -                               & -                               & 33.4        & 47.6 & 34.7            \\
\textbf{TGVAE} \cite{wang2019topic}                                & 48.73                           & 57.11                           & 87.86                           & -             & - & -          \\
\textbf{DSS-VAE} \cite{bao2019generating}                                 & -                               & -                               & -                               & 36.53        & - & -            \\
 {\textbf{TATGM} \cite{tang2019topic}}               &  {47.23}          &  {52.01}          &  {80.78}          &  {-}   & 40.80 & 32.90   \\
\textbf{iVAE} \cite{fang2019implicit}                                 & -                           & -                           & -                           & 53.44         & 47.93 & 36.88         \\
\textbf{VRTM} \cite{rezaee2020discrete}                                 & 47.78                           & 51.08                           & 86.33                           & 55.82        & - & -           \\ 
\textbf{Optimus} \cite{li2020optimus}                                 & -                           & -                           & -                           & 26.69         & 23.11 & 22.79     \\
\textbf{AdaVAE} \cite{tu2022adavae} & -                           & -                           & -                           & 11.97         & 14.23  & 15.49     \\
\textbf{DPrior} \cite{fang-etal-2022-controlled} & -                           & -                           & -                           & 14.74         & 14.67  & 14.52     \\
\textbf{DELLA} \cite{hu2022fuse} & -                           & -                           & -                           & -         & 11.49  & 12.35     \\
\bottomrule[1.5pt]
\end{tabular}
\caption{Text quality analysis in terms of text perplexity (\textit{PPL}) of LVMs on 6 common datasets. Results were gathered from the original papers, and ``-'' indicates no reported results from the original paper.}
\label{tab:ppl}
\end{table*}
\begin{itemize}
    \item \textbf{Reconstruction Loss} is the reconstruction loss (log-likelihood in general) of a language model. When the reconstruction loss is low, the model can generate authentic texts.
    \item \textbf{Kullback-Leibler divergence (KLD)} is the latent regularization metric for latent variable model. How to balance the KLD values is actually tricky. Because high KLD indicates that the model is not proper-trained and may not converge at all, while low KLD means there is likely the KL collapse issue happening and the model can simply degenerate to auto-encoder to copy training sentences for generation.
    \item \textbf{Evidence Lower Bound (ELBO)} is the sum of KL divergence and reconstruction loss (log-likelihood in general) of the latent variable model. The model reaches a better convergence when the ELBO is lower.
    \item \textbf{Activated Units (AU)} is used to measure the activated dimension in latent space. It is calculated by counting the number of posterior latent units that differ from the prior by a preassigned threshold. A higher AU value indicates that the model has a collapsed posterior.
    \item \textbf{Mutual Information (MI)} is the mutual information between input texts and latent posterior. It is generally implemented in the importance sampling method. When the MI between input and latent representations is high, the latent space learns more textual knowledge that may be beneficial for generation.
    \item \textbf{Perplexity (PPL)} is one of the main measurements for evaluating sequence fluency via word output probability from the model: for a given sequence of words $\{x_1, ..., x_N\}$ with $N$ to be the size, PPL is calculated as the normalized inverse probability of the sequence $PPL = p(x_1, ...x_N)^{-1/N}$. For latent variable models, there are two different ways to conduct PPL calculation. One is simply the negative exponential value of ELBO, which may be biased especially when the KL divergence does not collapse \cite{li2019surprisingly}. The other one applies importance weighted samples \cite{burda2015importance} to conduct an unbiased estimation of model PPL values. When the PPL is lower, the model is expected to produce more fluent contexts. We show PPL values of different controllable LVMs in Table \ref{tab:ppl}.
    \item \textbf{\textit{test}-BLEU} \cite{wang2019topic} is a generation metric related BLEU. It takes randomly sampled test examples as references, and computes the BLEU scores of generated texts with references. When \textit{test}-BLEU is high, the model produces more realistic-looking sentences.
    \item \textbf{\textit{self}-BLEU} \cite{zhu2018texygen} is a diversity metric related BLEU. It computes averaged BLEU scores among generated paragraphs, which randomly takes generated samples as references and others to compare with them. When \textit{self}-BLEU is low, the model can generate more diverse sentences.
    \item \textbf{Distinct-$n$} \cite{li2015diversity} is a diversity metric. It represents the number of distinct $n$-grams normalized by the length of text. The higher the Distinct scores are, the less likely the model produces ``dull texts''.
    \item \textbf{Human Evaluation} often takes \textit{fluency} and \textit{diversity} into consideration. The fluency score is high if the generated sentences are fluent, authentic, and correct w.r.t. common sense, while the diversity score is high when the generated texts are not dull and diverse in contextual structure, expression pattern, vocabulary, etc.
\end{itemize}
\subsection{Control Ability Metrics}
For text controllable generation task, there are multiple methods to verify the models' control degree:
\begin{itemize}
    \item \textbf{Accuracy on Token Level} measures the classification accuracy of generated controllable texts with given labels.
    \item \textbf{Accuracy on Latent Level} measures the classification accuracy of learned latent representations with given labels.
    \item \textbf{Normlized PMI (NPMI)} \cite{newman2010automatic} is a coherence metric of inferred topics from a model \footnote{\url{https://github.com/jhlau/topic-coherence-sensitivity}}. Given the top $n$ likely words of a topic, the coherence is calculated based on the sum of pairwise NPMI scores between topic words, where the word probabilities used in the NPMI calculation are based on co-occurrence statistics mined from English Wikipedia with a sliding window. When the NPMI value is higher, the topic inferred from the model is more concentrated and accurate.
    \item \textbf{Topic Entropy} \cite{rezaee2020discrete} is the entropy value of the latent representations regard to controlled topics, it helps us obtain the focused intensity of the topic modeling part with different documents. The lower entropy is, the fewer topics a topic model infers for one document, i.e., the higher the control level.
    \item \textbf{Keywords Extraction Accuracy} is specially designed for controllable text generation methods with assigned keywords inputs, i.e., hard constraints. It calculates the extraction accuracy of keywords from generated sentences compared with given ones.
    \item \textbf{Human Evaluation} often takes \textit{relevance} into consideration. When the generated sentences are correlated to given control signs, the evaluators tend to give higher relevance scores
\end{itemize}

\section{Datasets}
There are two types of datasets that are employed in controllable generation task. Datasets with label annotations or datasets without labels. Since vanilla latent variable models are famous for their knowledgeable latent representations, unsupervised latent variable models can conduct both textual feature extraction and generation tasks on unlabeled datasets such as APNEWS\footnote{\url{https://www.ap.org/en-gb/}}, IMDB \cite{maas2011learning}, BNC \cite{bnc2007british}, PTB \cite{marcus1993building}, SNLI \cite{bowman2015large}. 

For labeled datasets, there are globally labeled datasets (i.e., control on topic, sentiment, etc.) and keywords labeled datasets (i.e., control on specific keywords/phrases), we refer them to soft control and hard control datasets respectively.
\subsection{Soft Control Datasets}
The most widely used one is Yelp\footnote{\url{https://www.yelp.com/dataset}} dataset, which originally consists of examples from  Yelp restaurant reviews with ratings ranging from 1 to 5. \citet{shen2017style} processed this dataset with sentiment binary labels (rating above 3 as positive, otherwise negative), which becomes one of the universal versions to conduct controllable generation task. \citet{shen2020educating} further expanded the Yelp datasets into versions with text length labels (short, middle, long), text tense labels (past, present). Beyond the Yelp dataset, there are two more labeled corpora used in controllable LVMs. News Titles \cite{fu2018style} is a labeled dataset with \textit{Business}, \textit{Entertainment}, \textit{Health}, and \textit{Science} categories. Yahoo Question\footnote{\url{https://webscope.sandbox.yahoo.com}} dataset has 10 classes (\textit{Society}\&\textit{Culture}, \textit{Science}\&\textit{Mathematics}, \textit{Health}, \textit{Education}\&\textit{Reference}, \textit{Computers}\&\textit{Internet}, \textit{Sports}, \textit{Business}\&\textit{Finance}, \textit{Entertainment}\&\textit{Music}, \textit{Family}\&\textit{Relationships}, \textit{Politics}\&\textit{Government}.), each example has a question and long as well as short answers corresponding to the question.
\subsection{Hard Control Datasets}
Samples in hard control dataset generally consist of a sequence of key phrases and a fluent sentence that include all listed key phrases. \citet{shao2019long} proposed a Chinese dataset about shopping based on an online E-commerce website, they take clothing attributes as key phrases in the dataset. \textit{WiKiBio} dataset \cite{lebret2016neural} contains sentences of biographies from Wikipedia, they take different attributes of a celebrity such as a name, country, birth information, etc. \textit{SPNLG} dataset \cite{reed2018can} is a collection of restaurant descriptions, which expands the E2E dataset\footnote{\url{http://www.macs.hw.ac.uk/InteractionLab/E2E/}} with more varied sentence structures and instances \cite{ye2020variational}. 
\section{Conclusion}
\textit{The common misconception is that language has to do with words and what they mean. It doesn’t. It has to do with people and what they mean.} Controllable text generation is of vital importance to our daily applications. In order to decouple textual features to be controlled (e.g., topics, tense, sentiment) and syntax structure of basic grammar rules, deep latent variable models, which combine the composability and interpretability of graphical models with the flexible modeling capabilities of deep networks, are an exciting area of research. 

While it appears that a choice of one deep neural network over another is style-independent, balancing the trade-offs between the complexity of a model and the expected performance gains added by auxiliary components (e.g., classifier, discriminator) are consistent challenges faced by researchers. Since this overview is structured around the training paradigm of one model, for example, we can hardly find models to resolve table-to-text task which requires more specific control out of the supervised manner. Methods without external control signals have a more strict demand for latent awareness or well-designed control parts to conduct open-domain controllable generation. As the pre-trained language models arise, the semi-supervised manner is of great potential for future exploration and generalization.

It is our hope that this review would serve as an initial guideline for future studies that are built on the best practices of past research as well as some raw ideas that can enrich the field.
\newpage
\bibliography{mypaper} 

\begin{thebibliography}{98}
\expandafter\ifx\csname natexlab\endcsname\relax\def\natexlab#1{#1}\fi

\bibitem[{Bahdanau et~al.(2014)Bahdanau, Cho, and Bengio}]{bahdanau2014neural}
Dzmitry Bahdanau, Kyunghyun Cho, and Yoshua Bengio. 2014.
\newblock Neural machine translation by jointly learning to align and
  translate.
\newblock \emph{arXiv preprint arXiv:1409.0473}.

\bibitem[{Bao et~al.(2019)Bao, Zhou, Huang, Li, Mou, Vechtomova, Dai, and
  Chen}]{bao2019generating}
Yu~Bao, Hao Zhou, Shujian Huang, Lei Li, Lili Mou, Olga Vechtomova, Xinyu Dai,
  and Jiajun Chen. 2019.
\newblock Generating sentences from disentangled syntactic and semantic spaces.
\newblock \emph{arXiv preprint arXiv:1907.05789}.

\bibitem[{Bengio et~al.(2013)Bengio, L{\'e}onard, and
  Courville}]{bengio2013estimating}
Yoshua Bengio, Nicholas L{\'e}onard, and Aaron Courville. 2013.
\newblock Estimating or propagating gradients through stochastic neurons for
  conditional computation.
\newblock \emph{arXiv preprint arXiv:1308.3432}.

\bibitem[{Bowman et~al.(2015{\natexlab{a}})Bowman, Angeli, Potts, and
  Manning}]{bowman2015large}
Samuel~R Bowman, Gabor Angeli, Christopher Potts, and Christopher~D Manning.
  2015{\natexlab{a}}.
\newblock A large annotated corpus for learning natural language inference.
\newblock \emph{arXiv preprint arXiv:1508.05326}.

\bibitem[{Bowman et~al.(2015{\natexlab{b}})Bowman, Vilnis, Vinyals, Dai,
  Jozefowicz, and Bengio}]{bowman2015generating}
Samuel~R Bowman, Luke Vilnis, Oriol Vinyals, Andrew~M Dai, Rafal Jozefowicz,
  and Samy Bengio. 2015{\natexlab{b}}.
\newblock Generating sentences from a continuous space.
\newblock \emph{arXiv preprint arXiv:1511.06349}.

\bibitem[{Burda et~al.(2015)Burda, Grosse, and
  Salakhutdinov}]{burda2015importance}
Yuri Burda, Roger Grosse, and Ruslan Salakhutdinov. 2015.
\newblock Importance weighted autoencoders.
\newblock \emph{arXiv preprint arXiv:1509.00519}.

\bibitem[{Chen et~al.(2018)Chen, Li, Grosse, and Duvenaud}]{chen2018isolating}
Ricky~TQ Chen, Xuechen Li, Roger Grosse, and David Duvenaud. 2018.
\newblock Isolating sources of disentanglement in variational autoencoders.
\newblock \emph{arXiv preprint arXiv:1802.04942}.

\bibitem[{Cheng et~al.(2020)Cheng, Min, Shen, Malon, Zhang, Li, and
  Carin}]{cheng2020improving}
Pengyu Cheng, Martin~Renqiang Min, Dinghan Shen, Christopher Malon, Yizhe
  Zhang, Yitong Li, and Lawrence Carin. 2020.
\newblock Improving disentangled text representation learning with
  information-theoretic guidance.
\newblock \emph{arXiv preprint arXiv:2006.00693}.

\bibitem[{Consortium et~al.(2007)}]{bnc2007british}
BNC Consortium et~al. 2007.
\newblock British national corpus.
\newblock \emph{Oxford Text Archive Core Collection}.

\bibitem[{Dai et~al.(2019)Dai, Yang, Yang, Carbonell, Le, and
  Salakhutdinov}]{dai2019transformer}
Zihang Dai, Zhilin Yang, Yiming Yang, Jaime Carbonell, Quoc~V Le, and Ruslan
  Salakhutdinov. 2019.
\newblock Transformer-xl: Attentive language models beyond a fixed-length
  context.
\newblock \emph{arXiv preprint arXiv:1901.02860}.

\bibitem[{Dayan(2000)}]{dayan2000helmholtz}
Peter Dayan. 2000.
\newblock Helmholtz machines and wake-sleep learning.
\newblock \emph{Handbook of Brain Theory and Neural Network. MIT Press,
  Cambridge, MA}, 44(0).

\bibitem[{Devlin et~al.(2018)Devlin, Chang, Lee, and
  Toutanova}]{devlin2018bert}
Jacob Devlin, Ming-Wei Chang, Kenton Lee, and Kristina Toutanova. 2018.
\newblock Bert: Pre-training of deep bidirectional transformers for language
  understanding.
\newblock \emph{arXiv preprint arXiv:1810.04805}.

\bibitem[{Duan et~al.(2020)Duan, Xu, Pei, Han, and Li}]{duan2020pre}
Yuguang Duan, Canwen Xu, Jiaxin Pei, Jialong Han, and Chenliang Li. 2020.
\newblock Pre-train and plug-in: Flexible conditional text generation with
  variational auto-encoders.
\newblock In \emph{Proceedings of the 58th Annual Meeting of the Association
  for Computational Linguistics}, pages 253--262.

\bibitem[{Elhadad(1990)}]{elhadad1990constraint}
Michael Elhadad. 1990.
\newblock Constraint-based text generation using local constraints and
  argumentation to generate a turn in conversation.

\bibitem[{Fang et~al.(2019)Fang, Li, Gao, Dong, and Chen}]{fang2019implicit}
Le~Fang, Chunyuan Li, Jianfeng Gao, Wen Dong, and Changyou Chen. 2019.
\newblock Implicit deep latent variable models for text generation.
\newblock \emph{arXiv preprint arXiv:1908.11527}.

\bibitem[{Fang et~al.(2021)Fang, Zeng, Liu, Bo, Dong, and
  Chen}]{fang2021transformer}
Le~Fang, Tao Zeng, Chaochun Liu, Liefeng Bo, Wen Dong, and Changyou Chen. 2021.
\newblock Transformer-based conditional variational autoencoder for
  controllable story generation.
\newblock \emph{arXiv preprint arXiv:2101.00828}.

\bibitem[{Fang et~al.(2022)Fang, Li, Shang, Jiang, Liu, and
  Yeung}]{fang-etal-2022-controlled}
Xianghong Fang, Jian Li, Lifeng Shang, Xin Jiang, Qun Liu, and Dit-Yan Yeung.
  2022.
\newblock \href {https://aclanthology.org/2022.findings-acl.10} {Controlled
  text generation using dictionary prior in variational autoencoders}.
\newblock In \emph{Findings of the Association for Computational Linguistics:
  ACL 2022}, pages 97--111, Dublin, Ireland. Association for Computational
  Linguistics.

\bibitem[{Fu et~al.(2019)Fu, Li, Liu, Gao, Celikyilmaz, and
  Carin}]{fu2019cyclical}
Hao Fu, Chunyuan Li, Xiaodong Liu, Jianfeng Gao, Asli Celikyilmaz, and Lawrence
  Carin. 2019.
\newblock Cyclical annealing schedule: A simple approach to mitigating kl
  vanishing.
\newblock \emph{arXiv preprint arXiv:1903.10145}.

\bibitem[{Fu et~al.(2018)Fu, Tan, Peng, Zhao, and Yan}]{fu2018style}
Zhenxin Fu, Xiaoye Tan, Nanyun Peng, Dongyan Zhao, and Rui Yan. 2018.
\newblock Style transfer in text: Exploration and evaluation.
\newblock In \emph{Proceedings of the AAAI Conference on Artificial
  Intelligence}, volume~32.

\bibitem[{Garbacea and Mei(2020)}]{garbacea2020neural}
Cristina Garbacea and Qiaozhu Mei. 2020.
\newblock Neural language generation: Formulation, methods, and evaluation.
\newblock \emph{arXiv preprint arXiv:2007.15780}.

\bibitem[{Ghabussi et~al.(2019)Ghabussi, Mou, and
  Vechtomova}]{ghabussi2019stylized}
Amirpasha Ghabussi, Lili Mou, and Olga Vechtomova. 2019.
\newblock Stylized text generation using wasserstein autoencoders with a
  mixture of gaussian prior.
\newblock \emph{arXiv preprint arXiv:1911.03828}.

\bibitem[{Graves(2013)}]{graves2013generating}
Alex Graves. 2013.
\newblock Generating sequences with recurrent neural networks.
\newblock \emph{arXiv preprint arXiv:1308.0850}.

\bibitem[{Guo et~al.(2018)Guo, Lu, Cai, Zhang, Yu, and Wang}]{guo2018long}
Jiaxian Guo, Sidi Lu, Han Cai, Weinan Zhang, Yong Yu, and Jun Wang. 2018.
\newblock Long text generation via adversarial training with leaked
  information.
\newblock In \emph{Proceedings of the AAAI Conference on Artificial
  Intelligence}, volume~32.

\bibitem[{Higgins et~al.(2016)Higgins, Matthey, Pal, Burgess, Glorot,
  Botvinick, Mohamed, and Lerchner}]{higgins2016beta}
Irina Higgins, Loic Matthey, Arka Pal, Christopher Burgess, Xavier Glorot,
  Matthew Botvinick, Shakir Mohamed, and Alexander Lerchner. 2016.
\newblock beta-vae: Learning basic visual concepts with a constrained
  variational framework.

\bibitem[{Houlsby et~al.(2019)Houlsby, Giurgiu, Jastrzebski, Morrone,
  De~Laroussilhe, Gesmundo, Attariyan, and Gelly}]{houlsby2019parameter}
Neil Houlsby, Andrei Giurgiu, Stanislaw Jastrzebski, Bruna Morrone, Quentin
  De~Laroussilhe, Andrea Gesmundo, Mona Attariyan, and Sylvain Gelly. 2019.
\newblock Parameter-efficient transfer learning for nlp.
\newblock In \emph{International Conference on Machine Learning}, pages
  2790--2799. PMLR.

\bibitem[{Hu et~al.(2022)Hu, Yi, Li, Sun, and Xie}]{hu2022fuse}
Jinyi Hu, Xiaoyuan Yi, Wenhao Li, Maosong Sun, and Xing Xie. 2022.
\newblock Fuse it more deeply! a variational transformer with layer-wise latent
  variable inference for text generation.
\newblock In \emph{Proceedings of the 2022 Conference of the North American
  Chapter of the Association for Computational Linguistics: Human Language
  Technologies}, pages 697--716.

\bibitem[{Hu et~al.(1997)Hu, Palreddy, and Tompkins}]{hu1997patient}
Yu~Hen Hu, Surekha Palreddy, and Willis~J Tompkins. 1997.
\newblock A patient-adaptable ecg beat classifier using a mixture of experts
  approach.
\newblock \emph{IEEE transactions on biomedical engineering}, 44(9):891--900.

\bibitem[{Hu et~al.(2017)Hu, Yang, Liang, Salakhutdinov, and
  Xing}]{hu2017toward}
Zhiting Hu, Zichao Yang, Xiaodan Liang, Ruslan Salakhutdinov, and Eric~P Xing.
  2017.
\newblock Toward controlled generation of text.
\newblock In \emph{International Conference on Machine Learning}, pages
  1587--1596. PMLR.

\bibitem[{Iyyer et~al.(2014)Iyyer, Boyd-Graber, Claudino, Socher, and
  Daum{\'e}~III}]{iyyer2014neural}
Mohit Iyyer, Jordan Boyd-Graber, Leonardo Claudino, Richard Socher, and Hal
  Daum{\'e}~III. 2014.
\newblock A neural network for factoid question answering over paragraphs.
\newblock In \emph{Proceedings of the 2014 conference on empirical methods in
  natural language processing (EMNLP)}, pages 633--644.

\bibitem[{Kim et~al.(2018)Kim, Wiseman, and Rush}]{kim2018tutorial}
Yoon Kim, Sam Wiseman, and Alexander~M Rush. 2018.
\newblock A tutorial on deep latent variable models of natural language.
\newblock \emph{arXiv preprint arXiv:1812.06834}.

\bibitem[{Kingma et~al.(2014)Kingma, Mohamed, Rezende, and
  Welling}]{kingma2014semi}
Diederik~P Kingma, Shakir Mohamed, Danilo~Jimenez Rezende, and Max Welling.
  2014.
\newblock Semi-supervised learning with deep generative models.
\newblock In \emph{Advances in neural information processing systems}, pages
  3581--3589.

\bibitem[{Kingma and Welling(2013)}]{kingma2013auto}
Diederik~P Kingma and Max Welling. 2013.
\newblock Auto-encoding variational bayes.
\newblock \emph{arXiv preprint arXiv:1312.6114}.

\bibitem[{Kingma et~al.(2015)Kingma, Salimans, and
  Welling}]{kingma2015variational}
Durk~P Kingma, Tim Salimans, and Max Welling. 2015.
\newblock Variational dropout and the local reparameterization trick.
\newblock \emph{Advances in neural information processing systems}, 28.

\bibitem[{Lebret et~al.(2016)Lebret, Grangier, and Auli}]{lebret2016neural}
R{\'e}mi Lebret, David Grangier, and Michael Auli. 2016.
\newblock Neural text generation from structured data with application to the
  biography domain.
\newblock \emph{arXiv preprint arXiv:1603.07771}.

\bibitem[{Li et~al.(2019)Li, He, Neubig, Berg-Kirkpatrick, and
  Yang}]{li2019surprisingly}
Bohan Li, Junxian He, Graham Neubig, Taylor Berg-Kirkpatrick, and Yiming Yang.
  2019.
\newblock A surprisingly effective fix for deep latent variable modeling of
  text.
\newblock \emph{arXiv preprint arXiv:1909.00868}.

\bibitem[{Li et~al.(2020)Li, Gao, Li, Peng, Li, Zhang, and Gao}]{li2020optimus}
Chunyuan Li, Xiang Gao, Yuan Li, Baolin Peng, Xiujun Li, Yizhe Zhang, and
  Jianfeng Gao. 2020.
\newblock Optimus: Organizing sentences via pre-trained modeling of a latent
  space.
\newblock \emph{arXiv preprint arXiv:2004.04092}.

\bibitem[{Li et~al.(2015)Li, Galley, Brockett, Gao, and
  Dolan}]{li2015diversity}
Jiwei Li, Michel Galley, Chris Brockett, Jianfeng Gao, and Bill Dolan. 2015.
\newblock A diversity-promoting objective function for neural conversation
  models.
\newblock \emph{arXiv preprint arXiv:1510.03055}.

\bibitem[{Liu and Liu(2019)}]{liu2019transformer}
Danyang Liu and Gongshen Liu. 2019.
\newblock A transformer-based variational autoencoder for sentence generation.
\newblock In \emph{2019 International Joint Conference on Neural Networks
  (IJCNN)}, pages 1--7. IEEE.

\bibitem[{Liu and Lapata(2018)}]{liu2018learning}
Yang Liu and Mirella Lapata. 2018.
\newblock Learning structured text representations.
\newblock \emph{Transactions of the Association for Computational Linguistics},
  6:63--75.

\bibitem[{Logeswaran et~al.(2018)Logeswaran, Lee, and
  Bengio}]{logeswaran2018content}
Lajanugen Logeswaran, Honglak Lee, and Samy Bengio. 2018.
\newblock Content preserving text generation with attribute controls.
\newblock \emph{Advances in Neural Information Processing Systems}, 31.

\bibitem[{Maas et~al.(2011)Maas, Daly, Pham, Huang, Ng, and
  Potts}]{maas2011learning}
Andrew Maas, Raymond~E Daly, Peter~T Pham, Dan Huang, Andrew~Y Ng, and
  Christopher Potts. 2011.
\newblock Learning word vectors for sentiment analysis.
\newblock In \emph{Proceedings of the 49th annual meeting of the association
  for computational linguistics: Human language technologies}, pages 142--150.

\bibitem[{Mai and Henderson(2021)}]{mai2021bag}
Florian Mai and James Henderson. 2021.
\newblock Bag-of-vectors autoencoders for unsupervised conditional text
  generation.
\newblock \emph{arXiv preprint arXiv:2110.07002}.

\bibitem[{Mai et~al.(2020)Mai, Pappas, Montero, Smith, and
  Henderson}]{mai2020plug}
Florian Mai, Nikolaos Pappas, Ivan Montero, Noah~A Smith, and James Henderson.
  2020.
\newblock Plug and play autoencoders for conditional text generation.
\newblock \emph{arXiv preprint arXiv:2010.02983}.

\bibitem[{Makhzani et~al.(2015)Makhzani, Shlens, Jaitly, Goodfellow, and
  Frey}]{makhzani2015adversarial}
Alireza Makhzani, Jonathon Shlens, Navdeep Jaitly, Ian Goodfellow, and Brendan
  Frey. 2015.
\newblock Adversarial autoencoders.
\newblock \emph{arXiv preprint arXiv:1511.05644}.

\bibitem[{Marcus et~al.(1993)Marcus, Santorini, and
  Marcinkiewicz}]{marcus1993building}
Mitchell Marcus, Beatrice Santorini, and Mary~Ann Marcinkiewicz. 1993.
\newblock Building a large annotated corpus of english: The penn treebank.

\bibitem[{Mayfield et~al.(2019)Mayfield, Madaio, Prabhumoye, Gerritsen,
  McLaughlin, Dixon-Rom{\'a}n, and Black}]{mayfield2019equity}
Elijah Mayfield, Michael Madaio, Shrimai Prabhumoye, David Gerritsen, Brittany
  McLaughlin, Ezekiel Dixon-Rom{\'a}n, and Alan~W Black. 2019.
\newblock Equity beyond bias in language technologies for education.
\newblock In \emph{Proceedings of the Fourteenth Workshop on Innovative Use of
  NLP for Building Educational Applications}, pages 444--460.

\bibitem[{Meehan(1977)}]{meehan1977tale}
James~R Meehan. 1977.
\newblock Tale-spin, an interactive program that writes stories.
\newblock In \emph{Ijcai}, volume~77, page 9198.

\bibitem[{Mercatali and Freitas(2021)}]{mercatali2021disentangling}
Giangiacomo Mercatali and Andr{\'e} Freitas. 2021.
\newblock Disentangling generative factors in natural language with discrete
  variational autoencoders.
\newblock \emph{arXiv preprint arXiv:2109.07169}.

\bibitem[{Mikolov et~al.(2010)Mikolov, Karafi{\'a}t, Burget,
  {\v{C}}ernock{\`y}, and Khudanpur}]{mikolov2010recurrent}
Tom{\'a}{\v{s}} Mikolov, Martin Karafi{\'a}t, Luk{\'a}{\v{s}} Burget, Jan
  {\v{C}}ernock{\`y}, and Sanjeev Khudanpur. 2010.
\newblock Recurrent neural network based language model.
\newblock In \emph{Eleventh annual conference of the international speech
  communication association}.

\bibitem[{Mou et~al.(2016)Mou, Song, Yan, Li, Zhang, and Jin}]{mou2016sequence}
Lili Mou, Yiping Song, Rui Yan, Ge~Li, Lu~Zhang, and Zhi Jin. 2016.
\newblock Sequence to backward and forward sequences: A content-introducing
  approach to generative short-text conversation.
\newblock \emph{arXiv preprint arXiv:1607.00970}.

\bibitem[{Naesseth et~al.(2017)Naesseth, Ruiz, Linderman, and
  Blei}]{naesseth2017reparameterization}
Christian Naesseth, Francisco Ruiz, Scott Linderman, and David Blei. 2017.
\newblock Reparameterization gradients through acceptance-rejection sampling
  algorithms.
\newblock In \emph{Artificial Intelligence and Statistics}, pages 489--498.
  PMLR.

\bibitem[{Newman et~al.(2010)Newman, Lau, Grieser, and
  Baldwin}]{newman2010automatic}
David Newman, Jey~Han Lau, Karl Grieser, and Timothy Baldwin. 2010.
\newblock Automatic evaluation of topic coherence.
\newblock In \emph{Human language technologies: The 2010 annual conference of
  the North American chapter of the association for computational linguistics},
  pages 100--108.

\bibitem[{Oord et~al.(2017)Oord, Vinyals, and Kavukcuoglu}]{oord2017neural}
Aaron van~den Oord, Oriol Vinyals, and Koray Kavukcuoglu. 2017.
\newblock Neural discrete representation learning.
\newblock \emph{arXiv preprint arXiv:1711.00937}.

\bibitem[{Paris(2015)}]{paris2015user}
Cecile Paris. 2015.
\newblock \emph{User modelling in text generation}.
\newblock Bloomsbury Publishing.

\bibitem[{Park and Lee(2021)}]{park2021finetuning}
Seongmin Park and Jihwa Lee. 2021.
\newblock Finetuning pretrained transformers into variational autoencoders.
\newblock \emph{arXiv preprint arXiv:2108.02446}.

\bibitem[{Radford et~al.(2019)Radford, Wu, Child, Luan, Amodei, Sutskever
  et~al.}]{radford2019language}
Alec Radford, Jeffrey Wu, Rewon Child, David Luan, Dario Amodei, Ilya
  Sutskever, et~al. 2019.
\newblock Language models are unsupervised multitask learners.
\newblock \emph{OpenAI blog}, 1(8):9.

\bibitem[{Reed et~al.(2018)Reed, Oraby, and Walker}]{reed2018can}
Lena Reed, Shereen Oraby, and Marilyn Walker. 2018.
\newblock Can neural generators for dialogue learn sentence planning and
  discourse structuring?
\newblock \emph{arXiv preprint arXiv:1809.03015}.

\bibitem[{Rezaee and Ferraro(2020)}]{rezaee2020discrete}
Mehdi Rezaee and Francis Ferraro. 2020.
\newblock A discrete variational recurrent topic model without the
  reparametrization trick.
\newblock \emph{arXiv preprint arXiv:2010.12055}.

\bibitem[{Rezende and Mohamed(2015)}]{rezende2015variational}
Danilo Rezende and Shakir Mohamed. 2015.
\newblock Variational inference with normalizing flows.
\newblock In \emph{International conference on machine learning}, pages
  1530--1538. PMLR.

\bibitem[{Rezende et~al.(2014)Rezende, Mohamed, and
  Wierstra}]{rezende2014stochastic}
Danilo~Jimenez Rezende, Shakir Mohamed, and Daan Wierstra. 2014.
\newblock Stochastic backpropagation and approximate inference in deep
  generative models.
\newblock In \emph{International conference on machine learning}, pages
  1278--1286. PMLR.

\bibitem[{Ruiz et~al.(2016)Ruiz, AUEB, Blei et~al.}]{ruiz2016generalized}
Francisco~R Ruiz, Titsias~RC AUEB, David Blei, et~al. 2016.
\newblock The generalized reparameterization gradient.
\newblock \emph{Advances in neural information processing systems}, 29.

\bibitem[{Rush et~al.(2015)Rush, Chopra, and Weston}]{rush2015neural}
Alexander~M Rush, Sumit Chopra, and Jason Weston. 2015.
\newblock A neural attention model for abstractive sentence summarization.
\newblock \emph{arXiv preprint arXiv:1509.00685}.

\bibitem[{Semeniuta et~al.(2017)Semeniuta, Severyn, and
  Barth}]{semeniuta2017hybrid}
Stanislau Semeniuta, Aliaksei Severyn, and Erhardt Barth. 2017.
\newblock A hybrid convolutional variational autoencoder for text generation.
\newblock \emph{arXiv preprint arXiv:1702.02390}.

\bibitem[{Serban et~al.(2016)Serban, Sordoni, Bengio, Courville, and
  Pineau}]{serban2016building}
Iulian Serban, Alessandro Sordoni, Yoshua Bengio, Aaron Courville, and Joelle
  Pineau. 2016.
\newblock Building end-to-end dialogue systems using generative hierarchical
  neural network models.
\newblock In \emph{Proceedings of the AAAI Conference on Artificial
  Intelligence}, volume~30.

\bibitem[{Shao et~al.(2019)Shao, Huang, Wen, Xu, and Zhu}]{shao2019long}
Zhihong Shao, Minlie Huang, Jiangtao Wen, Wenfei Xu, and Xiaoyan Zhu. 2019.
\newblock Long and diverse text generation with planning-based hierarchical
  variational model.
\newblock \emph{arXiv preprint arXiv:1908.06605}.

\bibitem[{Shen et~al.(2017)Shen, Lei, Barzilay, and Jaakkola}]{shen2017style}
Tianxiao Shen, Tao Lei, Regina Barzilay, and Tommi Jaakkola. 2017.
\newblock Style transfer from non-parallel text by cross-alignment.
\newblock \emph{Advances in neural information processing systems}, 30.

\bibitem[{Shen et~al.(2020)Shen, Mueller, Barzilay, and
  Jaakkola}]{shen2020educating}
Tianxiao Shen, Jonas Mueller, Regina Barzilay, and Tommi Jaakkola. 2020.
\newblock Educating text autoencoders: Latent representation guidance via
  denoising.
\newblock In \emph{International Conference on Machine Learning}, pages
  8719--8729. PMLR.

\bibitem[{Shi et~al.(2020)Shi, Zhou, Miao, and Li}]{shi2020dispersed}
Wenxian Shi, Hao Zhou, Ning Miao, and Lei Li. 2020.
\newblock Dispersed exponential family mixture vaes for interpretable text
  generation.
\newblock In \emph{International Conference on Machine Learning}, pages
  8840--8851. PMLR.

\bibitem[{Sohn et~al.(2015)Sohn, Lee, and Yan}]{sohn2015learning}
Kihyuk Sohn, Honglak Lee, and Xinchen Yan. 2015.
\newblock Learning structured output representation using deep conditional
  generative models.
\newblock \emph{Advances in neural information processing systems},
  28:3483--3491.

\bibitem[{Subramanian et~al.(2018)Subramanian, Mudumba, Sordoni, Trischler,
  Courville, and Pal}]{subramanian2018towards}
Sandeep Subramanian, Sai~Rajeswar Mudumba, Alessandro Sordoni, Adam Trischler,
  Aaron~C Courville, and Chris Pal. 2018.
\newblock Towards text generation with adversarially learned neural outlines.
\newblock \emph{Advances in Neural Information Processing Systems}, 31.

\bibitem[{Tang et~al.(2019)Tang, Li, and Jin}]{tang2019topic}
Hongyin Tang, Miao Li, and Beihong Jin. 2019.
\newblock A topic augmented text generation model: Joint learning of semantics
  and structural features.
\newblock In \emph{Proceedings of the 2019 Conference on Empirical Methods in
  Natural Language Processing and the 9th International Joint Conference on
  Natural Language Processing (EMNLP-IJCNLP)}, pages 5090--5099.

\bibitem[{Tolstikhin et~al.(2017)Tolstikhin, Bousquet, Gelly, and
  Schoelkopf}]{tolstikhin2017wasserstein}
Ilya Tolstikhin, Olivier Bousquet, Sylvain Gelly, and Bernhard Schoelkopf.
  2017.
\newblock Wasserstein auto-encoders.
\newblock \emph{arXiv preprint arXiv:1711.01558}.

\bibitem[{Tomczak and Welling(2016)}]{tomczak2016improving}
Jakub~M Tomczak and Max Welling. 2016.
\newblock Improving variational auto-encoders using householder flow.
\newblock \emph{arXiv preprint arXiv:1611.09630}.

\bibitem[{Tu et~al.(2022{\natexlab{a}})Tu, Yang, Yang, Zhang, and
  Huang}]{tu2022adavae}
Haoqin Tu, Zhongliang Yang, Jinshuai Yang, Siyu Zhang, and Yongfeng Huang.
  2022{\natexlab{a}}.
\newblock Adavae: Exploring adaptive gpt-2s in variational auto-encoders for
  language modeling.
\newblock \emph{arXiv preprint arXiv:2205.05862}.

\bibitem[{Tu et~al.(2022{\natexlab{b}})Tu, Yang, Yang, Zhang, and
  Huang}]{tu2022pcae}
Haoqin Tu, Zhongliang Yang, Jinshuai Yang, Siyu Zhang, and Yongfeng Huang.
  2022{\natexlab{b}}.
\newblock Pcae: A framework of plug-in conditional auto-encoder for
  controllable text generation.
\newblock \emph{Knowledge-Based Systems}, page 109766.

\bibitem[{Turing(2009)}]{turing2009computing}
Alan~M Turing. 2009.
\newblock Computing machinery and intelligence.
\newblock In \emph{Parsing the turing test}, pages 23--65. Springer.

\bibitem[{Van~den Oord et~al.(2018)Van~den Oord, Li, and
  Vinyals}]{van2018representation}
Aaron Van~den Oord, Yazhe Li, and Oriol Vinyals. 2018.
\newblock Representation learning with contrastive predictive coding.
\newblock \emph{arXiv e-prints}, pages arXiv--1807.

\bibitem[{Vaswani et~al.(2017)Vaswani, Shazeer, Parmar, Uszkoreit, Jones,
  Gomez, Kaiser, and Polosukhin}]{vaswani2017attention}
Ashish Vaswani, Noam Shazeer, Niki Parmar, Jakob Uszkoreit, Llion Jones,
  Aidan~N Gomez, {\L}ukasz Kaiser, and Illia Polosukhin. 2017.
\newblock Attention is all you need.
\newblock In \emph{Advances in neural information processing systems}, pages
  5998--6008.

\bibitem[{Vincent et~al.(2010)Vincent, Larochelle, Lajoie, Bengio, Manzagol,
  and Bottou}]{vincent2010stacked}
Pascal Vincent, Hugo Larochelle, Isabelle Lajoie, Yoshua Bengio, Pierre-Antoine
  Manzagol, and L{\'e}on Bottou. 2010.
\newblock Stacked denoising autoencoders: Learning useful representations in a
  deep network with a local denoising criterion.
\newblock \emph{Journal of machine learning research}, 11(12).

\bibitem[{Vinyals et~al.(2015)Vinyals, Kaiser, Koo, Petrov, Sutskever, and
  Hinton}]{vinyals2015grammar}
Oriol Vinyals, {\L}ukasz Kaiser, Terry Koo, Slav Petrov, Ilya Sutskever, and
  Geoffrey Hinton. 2015.
\newblock Grammar as a foreign language.
\newblock \emph{Advances in neural information processing systems},
  28:2773--2781.

\bibitem[{Wang et~al.(2019{\natexlab{a}})Wang, Hua, and
  Wan}]{wang2019controllable}
Ke~Wang, Hang Hua, and Xiaojun Wan. 2019{\natexlab{a}}.
\newblock Controllable unsupervised text attribute transfer via editing
  entangled latent representation.
\newblock \emph{Advances in Neural Information Processing Systems},
  32:11036--11046.

\bibitem[{Wang et~al.(2018)Wang, Gan, Wang, Shen, Huang, Ping, Satheesh, and
  Carin}]{wang2018topic}
Wenlin Wang, Zhe Gan, Wenqi Wang, Dinghan Shen, Jiaji Huang, Wei Ping, Sanjeev
  Satheesh, and Lawrence Carin. 2018.
\newblock Topic compositional neural language model.
\newblock In \emph{International Conference on Artificial Intelligence and
  Statistics}, pages 356--365. PMLR.

\bibitem[{Wang et~al.(2019{\natexlab{b}})Wang, Gan, Xu, Zhang, Wang, Shen,
  Chen, and Carin}]{wang2019topic}
Wenlin Wang, Zhe Gan, Hongteng Xu, Ruiyi Zhang, Guoyin Wang, Dinghan Shen,
  Changyou Chen, and Lawrence Carin. 2019{\natexlab{b}}.
\newblock Topic-guided variational auto-encoder for text generation.
\newblock In \emph{NAACL-HLT (1)}.

\bibitem[{Wiseman et~al.(2017)Wiseman, Shieber, and
  Rush}]{wiseman2017challenges}
Sam Wiseman, Stuart~M Shieber, and Alexander~M Rush. 2017.
\newblock Challenges in data-to-document generation.
\newblock \emph{arXiv preprint arXiv:1707.08052}.

\bibitem[{Wiseman et~al.(2018)Wiseman, Shieber, and Rush}]{wiseman2018learning}
Sam Wiseman, Stuart~M Shieber, and Alexander~M Rush. 2018.
\newblock Learning neural templates for text generation.
\newblock \emph{arXiv preprint arXiv:1808.10122}.

\bibitem[{Xiao et~al.(2018)Xiao, Zhao, and Wang}]{xiao2018dirichlet}
Yijun Xiao, Tiancheng Zhao, and William~Yang Wang. 2018.
\newblock Dirichlet variational autoencoder for text modeling.
\newblock \emph{arXiv preprint arXiv:1811.00135}.

\bibitem[{Xu et~al.(2020)Xu, Cheung, and Cao}]{xu2020variational}
Peng Xu, Jackie Chi~Kit Cheung, and Yanshuai Cao. 2020.
\newblock On variational learning of controllable representations for text
  without supervision.
\newblock In \emph{International Conference on Machine Learning}, pages
  10534--10543. PMLR.

\bibitem[{Yang et~al.(2017)Yang, Hu, Salakhutdinov, and
  Berg-Kirkpatrick}]{yang2017improved}
Zichao Yang, Zhiting Hu, Ruslan Salakhutdinov, and Taylor Berg-Kirkpatrick.
  2017.
\newblock Improved variational autoencoders for text modeling using dilated
  convolutions.
\newblock In \emph{International conference on machine learning}, pages
  3881--3890. PMLR.

\bibitem[{Ye et~al.(2020)Ye, Shi, Zhou, Wei, and Li}]{ye2020variational}
Rong Ye, Wenxian Shi, Hao Zhou, Zhongyu Wei, and Lei Li. 2020.
\newblock Variational template machine for data-to-text generation.
\newblock \emph{arXiv preprint arXiv:2002.01127}.

\bibitem[{Yu and Koltun(2015)}]{yu2015multi}
Fisher Yu and Vladlen Koltun. 2015.
\newblock Multi-scale context aggregation by dilated convolutions.
\newblock \emph{arXiv preprint arXiv:1511.07122}.

\bibitem[{Yu et~al.(2017)Yu, Zhang, Wang, and Yu}]{yu2017seqgan}
Lantao Yu, Weinan Zhang, Jun Wang, and Yong Yu. 2017.
\newblock Seqgan: Sequence generative adversarial nets with policy gradient.
\newblock In \emph{Proceedings of the AAAI conference on artificial
  intelligence}, volume~31.

\bibitem[{Zhang et~al.(2017)Zhang, Gan, Fan, Chen, Henao, Shen, and
  Carin}]{zhang2017adversarial}
Yizhe Zhang, Zhe Gan, Kai Fan, Zhi Chen, Ricardo Henao, Dinghan Shen, and
  Lawrence Carin. 2017.
\newblock Adversarial feature matching for text generation.
\newblock In \emph{International Conference on Machine Learning}, pages
  4006--4015. PMLR.

\bibitem[{Zhang et~al.(2020)Zhang, Wang, Li, Gan, Brockett, and
  Dolan}]{zhang2020pointer}
Yizhe Zhang, Guoyin Wang, Chunyuan Li, Zhe Gan, Chris Brockett, and Bill Dolan.
  2020.
\newblock Pointer: Constrained progressive text generation via insertion-based
  generative pre-training.
\newblock \emph{arXiv preprint arXiv:2005.00558}.

\bibitem[{Zhang et~al.(2019)Zhang, Wang, Zhang, Zhang, and
  Gai}]{zhang2019improve}
Yuchi Zhang, Yongliang Wang, Liping Zhang, Zhiqiang Zhang, and Kun Gai. 2019.
\newblock Improve diverse text generation by self labeling conditional
  variational auto encoder.
\newblock In \emph{ICASSP 2019-2019 IEEE International Conference on Acoustics,
  Speech and Signal Processing (ICASSP)}, pages 2767--2771. IEEE.

\bibitem[{Zhao et~al.(2018)Zhao, Kim, Zhang, Rush, and
  LeCun}]{zhao2018adversarially}
Junbo Zhao, Yoon Kim, Kelly Zhang, Alexander Rush, and Yann LeCun. 2018.
\newblock Adversarially regularized autoencoders.
\newblock In \emph{International conference on machine learning}, pages
  5902--5911. PMLR.

\bibitem[{Zhao et~al.(2017{\natexlab{a}})Zhao, Song, and
  Ermon}]{zhao2017infovae}
Shengjia Zhao, Jiaming Song, and Stefano Ermon. 2017{\natexlab{a}}.
\newblock Infovae: Information maximizing variational autoencoders.
\newblock \emph{arXiv preprint arXiv:1706.02262}.

\bibitem[{Zhao et~al.(2017{\natexlab{b}})Zhao, Zhao, and
  Eskenazi}]{zhao2017learning}
Tiancheng Zhao, Ran Zhao, and Maxine Eskenazi. 2017{\natexlab{b}}.
\newblock Learning discourse-level diversity for neural dialog models using
  conditional variational autoencoders.
\newblock \emph{arXiv preprint arXiv:1703.10960}.

\bibitem[{Zhu et~al.(2018)Zhu, Lu, Zheng, Guo, Zhang, Wang, and
  Yu}]{zhu2018texygen}
Yaoming Zhu, Sidi Lu, Lei Zheng, Jiaxian Guo, Weinan Zhang, Jun Wang, and Yong
  Yu. 2018.
\newblock Texygen: A benchmarking platform for text generation models.
\newblock In \emph{The 41st International ACM SIGIR Conference on Research \&
  Development in Information Retrieval}, pages 1097--1100.

\end{thebibliography}

\end{document}